\documentclass[sigconf]{acmart}
\AtBeginDocument{%
  \providecommand\BibTeX{{%
    \normalfont B\kern-0.5em{\scshape i\kern-0.25em b}\kern-0.8em\TeX}}}

\copyrightyear{2023}
\acmYear{2023}
\setcopyright{acmlicensed}

\acmConference[KDD '23]{Proceedings of the 29th ACM SIGKDD Conference on Knowledge Discovery and Data Mining}{August 6--10, 2023}{Long Beach, CA, USA}
\acmBooktitle{Proceedings of the 29th ACM SIGKDD Conference on Knowledge Discovery and Data Mining (KDD '23), August 6--10, 2023, Long Beach, CA, USA}
\acmPrice{15.00}
\acmDOI{10.1145/3580305.3599510}
\acmISBN{979-8-4007-0103-0/23/08}

\usepackage{xspace} % self
\usepackage{amsmath,amsfonts} % self
\usepackage{bm} % self
\usepackage{bbm} % self
\usepackage{subfigure} % self
\usepackage{multirow} % self
\usepackage{booktabs} % self
\usepackage{enumitem} % self
\usepackage{array} % self
\usepackage{balance} % crv

\newcommand{\mymodel}{{SHGNN}\xspace}

\usepackage{xcolor}

\newcommand{\hide}[1]{}

\begin{document}

%%
%% The "title" command has an optional parameter,
%% allowing the author to define a "short title" to be used in page headers.
\title{Spatial Heterophily Aware  Graph Neural Networks}

\author{Congxi Xiao}
\authornote{This work was done when the first author was an intern at Baidu Research under the supervision of Jingbo Zhou.}
\affiliation{
  \institution{School of Computer Science and Technology,
  University of Science and Technology of China\\
  Baidu Research}
  \streetaddress{}
  \city{}
  \country{}
  }
\email{xiaocongxi@mail.ustc.edu.cn}
\orcid{0000-0001-5502-4788}

\author{Jingbo Zhou}
\authornote{Corresponding authors.}
\affiliation{
  \institution{Business Intelligence Lab,\\
  Baidu Research}
  \streetaddress{}
  \city{}
  \country{}
  }
\email{zhoujingbo@baidu.com}
\orcid{0000-0003-2677-7021}

\author{Jizhou Huang}
\affiliation{
  \institution{Baidu Inc.}
  \streetaddress{}
  \city{}
  \country{}
  }
\email{huangjizhou01@baidu.com}
\orcid{0000-0003-1022-0309}

\author{Tong Xu}
\affiliation{
  \institution{School of Computer Science and
Technology, University of Science and Technology of China\\
State Key Laboratory of Cognitive Intelligence}
  \streetaddress{}
  \city{}
  \country{}
  }
\email{tongxu@ustc.edu.cn}
\orcid{0000-0003-4246-5386}

\author{Hui Xiong}
\authornotemark[2]
\affiliation{
  \institution{
  The Thrust of Artificial Intelligence, The Hong Kong University of Science and Technology (Guangzhou), \\ The Department of Computer Science and Engineering, The Hong Kong University of Science and Technology}
  \streetaddress{}
  \city{}
  \country{}
  }
  \streetaddress{}
  \city{}
  \country{}
\email{xionghui@ust.hk}
\orcid{0000-0001-6016-6465}

%%
%% The "author" command and its associated commands are used to define
%% the authors and their affiliations.
%% Of note is the shared affiliation of the first two authors, and the
%% "authornote" and "authornotemark" commands
%% used to denote shared contribution to the research.

%%
%% By default, the full list of authors will be used in the page
%% headers. Often, this list is too long, and will overlap
%% other information printed in the page headers. This command allows
%% the author to define a more concise list
%% of authors' names for this purpose.
%\renewcommand{\shortauthors}{Xiao and Zhou, et al.}

\begin{abstract}
Graph Neural Networks (GNNs) have been broadly applied in many urban applications upon formulating a city as an urban graph whose nodes are urban objects like regions or points of interest.
Recently, a few enhanced GNN architectures have been developed to tackle heterophily graphs where connected nodes are dissimilar. However, urban graphs usually can be observed to possess a unique spatial heterophily property; that is, the dissimilarity of neighbors at different spatial distances can exhibit great diversity. This property has not been explored, while it often exists. To this end, in this paper, we propose a metric, named Spatial Diversity Score, to quantitatively measure the spatial heterophily and show how it can influence the performance of GNNs. Indeed, our experimental investigation clearly shows that existing heterophilic GNNs are still deficient in handling the urban graph with high spatial diversity score. This, in turn, may degrade their effectiveness in urban applications.
Along this line, we propose a Spatial Heterophily Aware Graph Neural Network (\mymodel), to tackle the spatial diversity of heterophily of urban graphs. Based on the key observation that spatially close neighbors on the urban graph present a more similar mode of difference to the central node, we first design a rotation-scaling spatial aggregation module, whose core idea is to properly group the spatially close neighbors and separately process each group with less diversity inside. Then, a heterophily-sensitive spatial interaction module is designed to adaptively capture the commonality and diverse dissimilarity in different spatial groups.
Extensive experiments on three real-world urban datasets demonstrate the superiority of our \mymodel over several its competitors.
\end{abstract}

%%
%% The code below is generated by the tool at http://dl.acm.org/ccs.cfm.
%% Please copy and paste the code instead of the example below.
%%


%%
%% Keywords. The author(s) should pick words that accurately describe
%% the work being presented. Separate the keywords with commas.

% \received{20 February 2007}
% \received[revised]{12 March 2009}
% \received[accepted]{5 June 2009}

\maketitle

\section{Introduction}
Applying Graph Neural Networks (GNNs) in different urban applications has attracted much research attention in the past few years \cite{song2020spatial, zhou2021competitive, fang2021spatial,xia20213dgcn, rao2022fogs,han2022semi,wu2020learning}.
These studies usually model the city as an urban graph whose nodes are urban objects (e.g., regions or Points of Interest (POIs)) and whose edges are  physical or social dependencies in the  urban area (e.g., human mobility and road connection \cite{wu2022multi,wu2020learning}). Upon urban graphs, GNNs with variant architectures are proposed to achieve the classification or regression tasks. 

\begin{figure*}[t]
\centering
\includegraphics[width=0.91\textwidth]{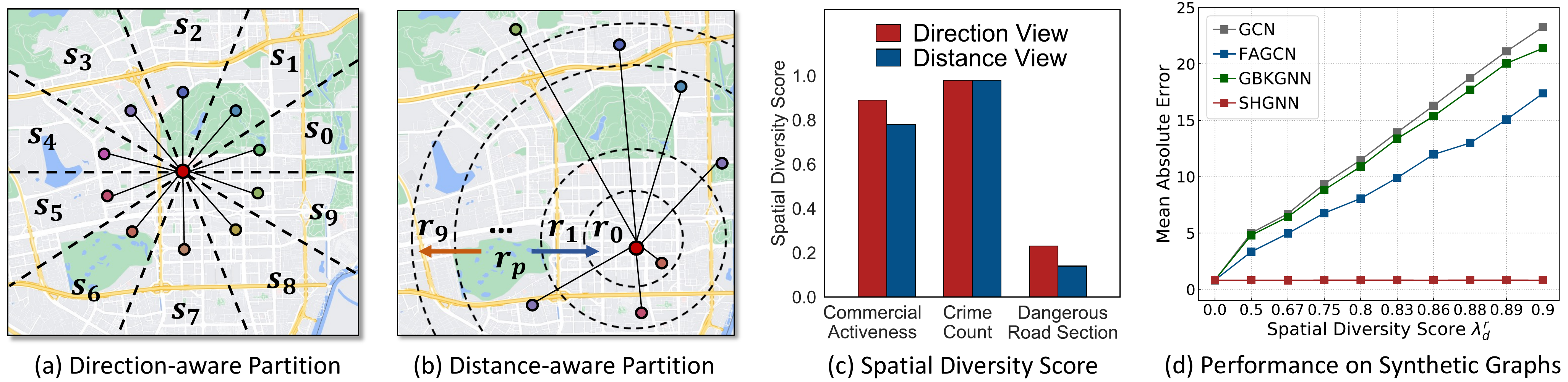}
\vspace{-3.5mm}
\caption{Analysis of spatial heterophily. (a)-(b) illustrate the space partition from two spatial views. (c) shows the spatial diversity scores calculated on three real-world urban graphs. (d) presents the results of experimental investigation.}
\label{fig_motivation}
\vspace{-4mm}
\end{figure*}

Nevertheless, there is a serious limitation of GNNs which has been largely overlooked in previous studies but have attracted increasing research attention recently: GNNs have an implicit homophily assumption that nodes only with similar features
or same labels are connected together on the graph \cite{zhu2021graph,zheng2022graph}. Meanwhile, the opposite assumption is heterophily that connected nodes have dissimilar features or labels. 
In fact, heterophily usually exists in an urban graph as it describes the complex urban system where both similar and dissimilar urban objects
(e.g., regions with different functionalities) 
can correlate with each other in complex manners.
Taking the urban graph constructed with human mobility as an example, the start node and end node of an edge could be home and workplace, respectively, which are definitely heterophilic.
Such difference information on the heterophilic urban graph may not be modeled well by many traditional homophilic GNNs, who tend to generate similar representations for connected nodes \cite{zhu2021graph, bo2021beyond}. In this way, the performance of these homophilic GNN methods on urban graphs may be largely hindered.

Our further observation is that urban graphs have a unique \emph{Spatial Heterophily} property. To be specific, we find that the heterophily on the urban graph often presents a characteristic of \emph{spatial diversity}. In other words, the difference (or dissimilarity) between the central node and its neighbors \hide{located }at different distances or directions exhibits evident discrepancy, rather than distributed uniformly. 
Aware of such a characteristic, an attendant question is how to measure the spatial heterophily of the urban graph.
There have been various studies put forward to investigate the graph homophily and heterophily from different perspectives, including node homophily \cite{pei2019geom}, edge homophily \cite{zhu2020beyond} and class homophily \cite{lim2021large}. But without considering the spatial position of linking nodes, these metrics cannot describe the spatial heterophily on urban graphs.

Therefore, in this work, we propose a metric named \textbf{spatial diversity score} to analyze the spatial heterophily, and investigate its influence on the performance of existing GNN
methods on urban graphs. Firstly, we divide the neighbors on an urban graph into different spatial groups according to their locations (including direction and distance), and then the spatial diversity score measures the discrepancy
between different spatial groups, in terms of their label dissimilarity to the central node. A higher score (close to one) indicates a larger discrepancy between different spatial groups, and thus a higher spatial diversity of heterophily on the urban graph.

Yet, it still remains an outstanding challenge for designing powerful heterophilic GNN models over an urban graph if its spatial diversity score is high, where there is a diversity of dissimilarity distributions between the central node and its neighbors at different spatial locations.
There are some recent studies to improve the GNN architectures to handle the graph heterophily \cite{jia2020residual,lim2021new, yang2021graph, kim2020find, du2022gbk}.
Most of these methods can only work on a heterophilic graph when there is limited difference between nodes. For example, GBKGNN \cite{du2022gbk} assumes that there are only two different kinds of nodes, and FAGCN \cite{bo2021beyond} assumes that node features have only two different levels of frequencies (Note that this limitation is discussed in their papers). Following this line, it is hard to model such diverse distributions of spatial heterophily on urban graphs.
To provide more evidence, we conduct an experiment on synthetic urban graphs with varying levels of spatial diversity score (details are in Section \ref{toy}). As shown in Figure \ref{fig_motivation}(d), when the graph presents higher spatial diversity score,\hide{of heterophily,} the performance of these two state-of-the-art heterophilic GNNs, GBKGNN and FAGCN, is far from optimal.
We also apply the proposed spatial diversity score to analyze three real-world urban graphs under different target tasks in our experiments. As we can see from Figure \ref{fig_motivation}(c), three urban graphs present different levels of spatial heterophily, where one of them can get a very high score (0.99 on the urban graph in crime prediction task). Thus, it is valuable to develop an effective GNN model that can handle the diverse spatial heterophily of the urban graph.

Through our in-depth analysis of spatial heterophily, we observe that the heterophily further exhibits a spatial tendency on urban graphs, which reveals a promising opportunity for us to tackle such diverse heterophily in a divide-and-conquer way. 
Different from ordinary graphs, nodes on urban graphs should follow Tobler's First Law of Geography (TFL) \cite{tobler1970computer}. As the fundamental assumption used in almost all urban analysis, TFL means \emph{everything is related to everything else, but near things are more related than distant things.} 
Obeying TFL, spatially close neighbors on the urban graph present a more similar mode of difference to the central node, compared to the distant ones.
We also analyze the real-world urban graph in commercial activeness prediction task to visualize such a tendency (details are in Section \ref{spatial_tendency_method}). Figure \ref{data_analysis} clearly shows that spatially close neighbors present less discrepancy from both the direction and distance view.
Thus, if we can properly
group the spatially close neighbors together, it is possible to alleviate the diversity of heterophily inside groups on the urban graph.

To this end, we propose a novel \underline{S}patial \underline{H}eterophily Aware \underline{G}raph \underline{N}eural \underline{N}etwork (\mymodel), to tackle the spatial heterophily on urban graphs, with two specially designed modules. 
First, we devise a Rotation-Scaling Spatial Aggregation module. Its core idea is to properly divide the neighbors into different spatial groups according to their direction and distance to the central node, and perform a spatial-aware feature aggregation for each group, which serves as the basis of handling diverse heterophily distributions separately.
Then, a Heterophily-Sensitive Spatial Interaction module with two learnable kernel functions is designed to capture the commonality and discrepancy in the neighborhood, and adaptively determine what and how much difference information the central node needs.
It acts between the central node and neighbors in different groups to manage the spatial diversity of heterophily on the urban graph.

The contribution of this paper is summarized as follows:
\begin{itemize}[leftmargin=20pt, topsep=2pt]
    \setlength{\itemsep}{0pt}
    \setlength{\parsep}{0pt}
    \setlength{\parskip}{0pt}
    \item To the best of our knowledge, we are the first to investigate the spatial heterophily of urban graphs. 
    We design a metric named spatial diversity score, to analyze the spatial heterophily property, and identify the limitation of existing GNNs in handling the diverse spatial heterophily on the urban graph.
    \item We propose a novel spatial heterophily aware graph neural network named \mymodel, in which two techniques: rotation-scaling spatial aggregation and heterophily-sensitive spatial interaction are devised to tackle the spatial heterophily of the urban graph in a divide-and-conquer way. 
    \item We conducted extensive experiments to verify the effectiveness of \mymodel on three real-world datasets.
    
    \hide{on three real-world datasets to demonstrate the effectiveness of our \mymodel across broad urban applications.}
\end{itemize}

\section{Preliminaries}
\label{pre}
In this section, we first introduce the basic concepts of the urban graph, then clarify the goal of our work. The frequently used notations are summarized in Table \ref{table-symbol} in Appendix.
\vspace{-1.5mm}
\paragraph{Urban Graphs.} 
Let $\mathcal{G}(\mathcal{V},\mathcal{E},\bm{X})$ denote an urban graph, where $\mathcal{V}=\{v_1, ..., v_N\}$ denotes a set of nodes representing a kind of urban entity, $\mathcal{E}$ denotes the edge set indicating one type of relation among nodes in the urban scenario, and  $\mathcal{N}(v_i) = \{v_j|(v_i, v_j)\in \mathcal{E}\}$ is the neighborhood of node $v_i$.
$\bm{X} \in \mathbb{R}^{N \times d}$ denotes the feature matrix, in which the $i$-th row is the $d$-dimensional node features of $v_i$ obtained from the urban data.
Different instantiations of node set and edge set will form different urban graphs, such as:
(1) \textbf{Mobility Graph} with regions as nodes and human flows as edges. The node features can be some region attributes such as the distribution of POI inside the region;
(2) \textbf{Road Network} where the node set is formed by road sections, and the edges denote the connectivity between them. The node features can be the structural information of a section, such as the number of branches and lanes.

\vspace{-1.5mm}
\paragraph{Problem Formulation.}
Given an urban graph, our goal is to design a GNN model, that considers and alleviates the spatial heterophily, to learn the node representation $f\!:(v_i | \mathcal{G}) \!\rightarrow\! \hat{\bm{h}}_i$,
where $\hat{\bm{h}}_i$ denotes the representation vector of node $v_i$. The model $f$ will be trained in an end-to-end manner in different downstream tasks.

\section{Spatial Heterophily}
In this section, an analysis of spatial heterophily on urban graphs is provided. 
We first introduce our metric to measure the spatial heterophily (Section \ref{discrepancy_metric}).
Then, Section \ref{toy} gives an experimental investigation on synthetic graphs. It not only demonstrates the importance for GNNs to consider the spatial heterophily on urban graphs, but also suggests a promising way to tackle this challenge.

\vspace{-1mm}
\subsection{Spatial Diversity of Heterophily}
\label{discrepancy_metric}

To describe the spatial diversity of heterophily on urban graphs, we design a metric named \textbf{spatial diversity score}.
Typically, the graph heterophily is measured by the label dissimilarity between the central node and its neighbors (e.g., \cite{pei2019geom,du2022gbk}). Following this line, our spatial diversity score aims to further assess the discrepancy between neighbors at different spatial locations, in terms of the distributions of their label dissimilarity to the central node.
Briefly, we first divide the neighborhood into different spatial groups according to their spatial locations. Then, we can measure spatial groups' discrepancy by calculating the Wasserstein distance between their label dissimilarity distributions, and the metric is further defined as the ratio of nodes with high discrepancy.

\vspace{-1mm}
\subsubsection{Dual-View Space Partition}
\label{metric_space_partition}
To distinguish the spatial location of neighbors and form different spatial groups, we first partition the geographic space into several non-overlap subspaces,
% from direction and distance view, 
and each neighbor of the central node can be assigned to the group corresponding to the subspace it locates in. 
Note that the spatial heterophily can both present in different directions and distances, thus we propose to perform space partition from both two views.

\textbf{Direction-Aware Partition.}
Given the central node $v_i$ on an urban graph, we evenly partition the geographic space centered by it into ten direction sectors $\mathcal{S} = \{s_k \, | \, k=0,1,...,9\}$.
Correspondingly, nodes in the neighborhood $\mathcal{N}(v_i)$ will be divided into the sector they locate in.
Neighbors belonging to the same sector are redefined as the direction-aware neighborhood $\{\mathcal{N}_{s_k}(v_i) \,|\, k=0,1,...,9\}$, where $\bigcup_{k=0}^{9} \mathcal{N}_{s_k}(v_i) \!=\! \mathcal{N}(v_i)$.
In this way, the direction-aware neighborhoods can be regarded as different spatial groups associated with different spatial relations to the central node. We will then calculate the discrepancy between different spatial groups. Figure \ref{fig_motivation}(a) illustrates such a sector partition.

\textbf{Distance-Aware Partition.}
As illustrated in Figure \ref{fig_motivation}(b), we also divide the neighbors based on their distance to the central node. 
To be specific, we first determine the distance range of the neighborhood on an urban graph, by making a statistic of the distance distribution between connected nodes. Note that we consider the $90\%$ percentile of this distribution as the maximum distance of the neighborhood on the graph. This is based on the observation that the distance distribution often presents a long tail property, and such a distance cut-off can avoid interference from extremely distant outliers.
And then, the distance range is evenly split into ten buckets, which results in distance rings $\mathcal{R} = \{r_k \, | \, k=0,1,...,9\}$. Similarly, the original neighborhood $\mathcal{N}(v_i)$ can be divided into these ten distance-aware neighborhoods $\{\mathcal{N}_{r_k}(v_i) \, | \, k=0,1,...,9\}$ as another view of spatial groups, where we also have $\bigcup_{k=0}^{9} \mathcal{N}_{r_k}(v_i) = \mathcal{N}(v_i)$.

\begin{figure*}[t]
\centering
\includegraphics[width=0.92\textwidth]{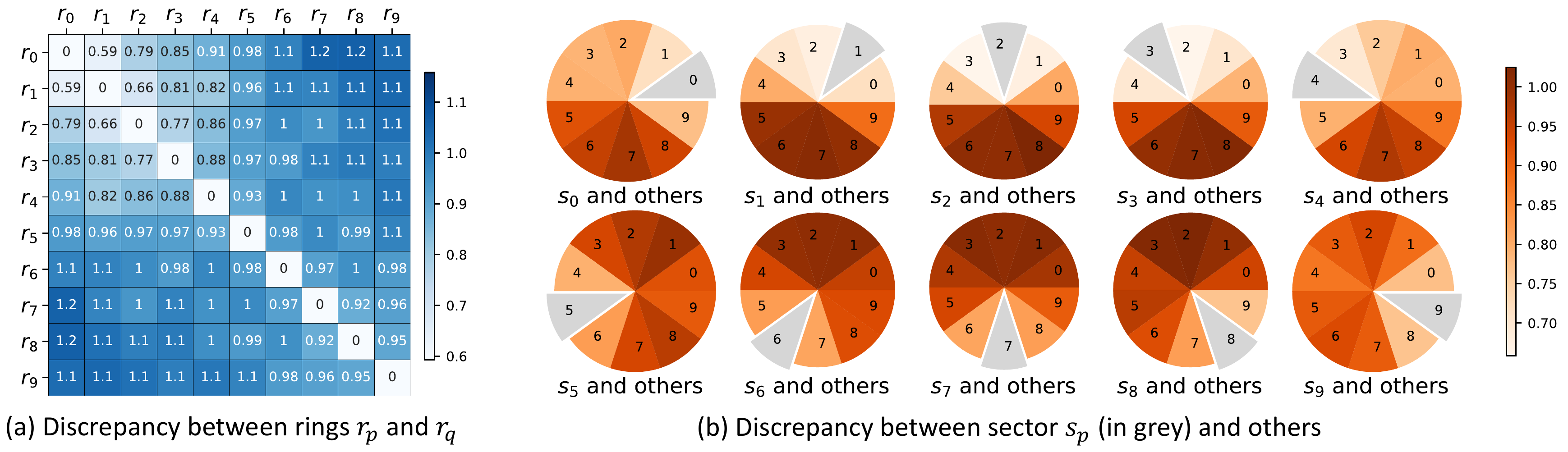}
 \vspace{-4mm}
\caption{Illustration of spatial tendency in the distance view (a) and direction view (b). 
}
\label{data_analysis}
\vspace{-4mm}
\end{figure*}

\subsubsection{Spatial Diversity Score}
After the neighborhood partition, our goal is to further define the spatial diversity score by measuring the discrepancy between different spatial groups, based on their distribution distance of label dissimilarity to the central node.

First of all, we define spatial group's label dissimilarity distribution to the central node. 
For a node classification task with $\mathcal{C}$ classes, within a spatial group, this distribution is calculated as the ratio of neighbors belonging to each class (different to the central node's).
Taking the direction view as an example, for the central node $v_i$, the label dissimilarity distribution of $s_k$ is formally defined as $P_i^{s_k} = [\, P_{i,0}^{s_k} \,,\, P_{i,1}^{s_k} \,,\, ... \,,\, P_{i,|\mathcal{C}-1|}^{s_k}]$ with:
\begin{equation}
    P_{i,c}^{s_k} = \sum_{v_j \in \mathcal{N}_{s_k}(v_i) \land y_j \neq y_i } \!\!\! \mathbbm{1}(y_{j,c} = 1) \cdot |\mathcal{N}_{s_k}(v_i)|^{-1},
\vspace{-1mm}
\end{equation}
where $c=0,1,...,|\mathcal{C}-1|$, $\mathbbm{1}(\cdot)$ is the indicator function, and $y_i$ is $v_i$'s one-hot label vector whose $c$-th value is denoted by $y_{i,c}$. 

In addition, to improve the generality of this metric to more urban applications (e.g., regression tasks), we also extend the definition of label dissimilarity distribution above to the node regression task.
To be specific, we first make a statistic of the label difference between connected nodes on the whole graph $\hat{\mathcal{Y}} = \{ (y_j-y_i) \,|\, (v_i, v_j) \in \mathcal{E} \}$, and calculate the deciles $\{D_1, D_2, ,..., D_9 \}$ of this distribution. These nine deciles can determine ten buckets (intervals), which will be used for the discretization of continuous label difference value, to obtain a similar form of label dissimilarity as the node classification task.
Hence, still for the spatial group $\mathcal{N}_{s_k}(v_i)$ in sector $s_k$, its label dissimilarity distribution will be calculated as the ratio of neighbors mapped into different buckets, according to their discretized label difference to the central node, which can be also formulized as:
$P_i^{s_k} = [\, P_{i,0}^{s_k} \,,\, P_{i,1}^{s_k} \,,\, ... \,,\, P_{i,9}^{s_k}]$, with the $c$-th element computed by:
\begin{equation}
    P_{i,c}^{s_k} = \sum_{v_j \in \mathcal{N}_{s_k}(v_i)} \!\!\! \mathbbm{1}(D_c < y_j - y_i \leq D_{c+1}) \cdot |\mathcal{N}_{s_k}(v_i)|^{-1},
\vspace{-1mm}
\end{equation}
where $c=0,1,...,9$. The $D_0$ and $D_{10}$ are used to denote the minimum and maximum of the label difference distribution on the whole graph, respectively (i.e., $D_0=\min(\hat{\mathcal{Y}})$ and $D_{10}=\max(\hat{\mathcal{Y}})$).

Next, the discrepancy between different spatial groups can be defined by measuring the distance between their label dissimilarity distributions. Following a recent study on the graph heterophily \cite{zhao2022neighborhood}, we adopt Wasserstein distance (WD) to measure the distribution distance between two spatial groups. 
Formally, consider two spatial groups $\mathcal{N}_{s_p}(v_i)$ and $\mathcal{N}_{s_q}(v_i)$ in sector $s_p$ and $s_q$ of node $v_i$, the discrepancy between them is defined as:
\begin{equation}
\label{disc_eq}
    Disc(v_i, s_p, s_q) = WD(P_i^{s_p}, P_i^{s_q}) ,
\end{equation}
where $WD(\cdot, \cdot)$ denotes the Wasserstein distance between two distributions, which can be approximately calculated by the Sinkhorn iteration algorithm \cite{cuturi2013sinkhorn}.

With such a measurement, we can finally define the spatial diversity score to describe the diverse spatial heterophily of an urban graph. This metric can be computed based on the ratio of nodes with high discrepancy among different spatial groups:
\begin{equation}
    \lambda_d^{s} = |\mathcal{V}|^{-1} \cdot \!\! \sum_{v_i \in \mathcal{V}} \mathbbm{1}(\max_{p \neq q}Disc(v_i, s_p, s_q) \geq 1) ,
\vspace{-0.5mm}
\end{equation}
where $\max_{p \neq q}Disc(v_i, s_p, s_q) \geq 1$ with $p, q=0,1,...,9$ indicates that there are at least two sectors being discrepant in terms of their label dissimilarity distributions to the central node $v_i$.
In this way, the score $\lambda_d^{s}$ will get higher if there are more nodes whose spatial groups present high discrepancy on the urban graph.
Similarly, we can also define the score $\lambda_d^{r}$ in the distance view, which measures the discrepancy between spatial groups formed by different rings in $\mathcal{R}$, with different distances to the central node:
\begin{equation}
    \lambda_d^{r} = |\mathcal{V}|^{-1} \cdot \!\! \sum_{v_i \in \mathcal{V}} \mathbbm{1}(\max_{p \neq q}Disc(v_i, r_p, r_q) \geq 1) ,
\vspace{-0.5mm}
\end{equation}

In practice, when only a part of nodes are labeled on an urban graph, it's often sufficient to use the labeled data to estimate $\lambda_d^{s}$ and $\lambda_d^{r}$.
Figure \ref{fig_motivation}(c) shows the scores of three real-world urban graphs.
As we can see, from both the direction and distance view, 
urban graphs can present very high spatial diversity scores (e.g., 0.99 in crime prediction), which reveals the diverse heterophily in different spatial groups, with discrepant label dissimilarity distributions.

\vspace{-1mm}
\subsection{Experimental Investigation}
\label{toy}
Next, we conduct an experimental investigation on synthetic urban graphs to illustrate the importance of considering the diverse spatial heterophily on the urban graph.
% and examine the effectiveness of neighborhood spatial grouping inspired by the property of spatial tendency on urban graphs.
Specifically, we test the performance of several GNN models on a series of synthetic urban graphs with increasing spatial diversity of the heterophily.
We generate 10 graphs containing 5000 nodes with 10-dimensional randomly generated feature vectors.
For each node, we build 50 edges and assume that these neighbors locate at 10 distance rings from near to far around the central node.
To increase the spatial diversity of heterophily from $\mathcal{G}_1$ to $\mathcal{G}_{10}$, we gradually enlarge the discrepancy of label dissimilarity distributions between neighbors in different distance rings.
Specifically, in graph $\mathcal{G}_i$, we evenly divide the node set into $i$ subsets $\mathcal{V}_i = \bigcup_{j=1}^{i} \mathcal{V}_{i,j}$, where the labels of nodes in $\mathcal{V}_{i,j}$ are sampled from the Gaussian distribution $\mathcal{N}(10j,1)$. 
Then, the 50 neighbors connected to the node in $\mathcal{V}_{i,j}$ are randomly selected from one of subsets $\{ \mathcal{V}_{i,k} \} |_{k=j}^{i}$ by an equal probability $1/(i-j+1)$. Besides, we let neighbors' spatial distance be consistent with their label difference to the central node (i.e., neighbors with smaller differences locates in the closer distance ring).
In this way, these 10 graphs will have increasing spatial diversity scores $\lambda_d^r$.
Meanwhile, neighbors in the same distance ring have less discrepancy.

Figure \ref{fig_motivation}(d) shows the node regression error of GCN \cite{kipf2016semi} and two state-of-the-art heterophic GNNs (FAGCN \cite{bo2021beyond} and GBKGNN \cite{du2022gbk}) on 10 synthetic graphs. As expected, GCN gets the worst performance, since it proved unsuitable for heterophily graphs in most cases \cite{lim2021new}. For two heterophic GNNs, despite less error than GCN, their performance is still far from optimal when the spatial diversity score increases. This is because they only model limited dissimilarity between nodes (e.g., two levels of frequencies \cite{bo2021beyond} or two classes of nodes \cite{du2022gbk}), without considering diverse dissimilarity distributions.
However, if the model separately processes neighbors in each distance ring whose differences to the central node are more similar, (as our \mymodel did), it can always achieve satisfactory performance.
Thus, it's reasonable to consider whether we can tackle the diverse spatial heterophily, by properly grouping together the neighbors with less discrepancy on the urban graph.

\vspace{-1mm}
\section{Methodology}
In this section, we first present that the heterophily on urban graphs often exhibits a spatial tendency: spatially close neighbors have more similar heterophily distributions than distant ones, which obeys TFL. This characteristic gives rise to the solution that we can properly divide neighbors according to their spatial locations, to achieve grouping neighbors with less discrepancy together.
Then, we introduce our proposed \mymodel that leverages such a spatial tendency to tackle the spatial heterophily of urban graphs.

\textbf{Spatial Tendency.}
\label{spatial_tendency_method}
In addition to the presence of a discrepancy between different spatial groups (discussed in Section \ref{discrepancy_metric}), our in-depth investigation of spatial heterophily further reveals that such a discrepancy shows a spatial tendency on urban graphs. 
Specifically, we observe that the discrepancy
between two spatially close groups is smaller, compared to two distant groups. 
We conduct a data analysis to visually present such a tendency.
Figure \ref{data_analysis} shows the pair-wise discrepancy of label dissimilarity distributions between any two spatial groups, which is computed based on real-world human mobility and regional commercial activeness data.
In Figure \ref{data_analysis}(a), for the spatial group formed by \hide{neighbors in }distance ring $r_p$, we can find that its discrepancy with other rings $r_q$ is highly correlated to their spatial distance (i.e., the discrepancy gets higher when $r_q$ is more distant to $r_p$). For example, \hide{the label dissimilarity distributions of }the close ring pair $(r_0,r_1)$ are less discrepant than the distant pair $(r_0,r_9)$.
A similar spatial tendency can also be observed from the direction view. 
As shown in Figure \ref{data_analysis}(b), in most cases, the discrepancy will increase along with the included angle between sectors. Taking sector $s_0$ as an example, its discrepancy with $s_1\sim s_9$ increases first and then decreases, which is roughly in sync with the change of their included angles.
In other words, a sector is more likely to be discrepant to another distant sector (e.g., $(s_1, s_6)$) than a nearby one (e.g., $(s_1, s_2)$).
This characteristic motivates us to address the diverse spatial heterophily on urban graphs, by properly grouping spatially close neighbors and separately processing each group with less discrepancy inside.

To this end, we propose a novel GNN architecture named \mymodel, which is illustrated in Figure \ref{fig_multi_head} and \ref{fig_framework}. Our model consists of two components: Rotation-Scaling Spatial Aggregation (see Section \ref{rot_sca_agg}) and Heterophily-Sensitive Spatial Interaction (see Section \ref{heter_aware}). 

\vspace{-1mm}
\subsection{Rotation-Scaling Spatial Aggregation}
\label{rot_sca_agg}

This component aims to properly group spatially close neighbors and alleviate the diversity of heterophily inside groups in the message passing process. In general, we first divide neighbors according to their relative positions to the central node. Then, the feature aggregation is performed in each spatial group separately. 

\subsubsection{Rotation-Scaling Dual-View Partition.}
Following the neighborhood partition in Section \ref{metric_space_partition}, we also partition the geographic space into non-overlap subspaces, and neighbors located at the same subspace are then grouped together. Note that there are two major differences between the space partition performed in this component and that in Section \ref{metric_space_partition}. First, we apply a more general partition with a variable number of subspaces. Second, we introduce a rotation-scaling multi-head partition strategy to model the neighbor's spatial location in a more comprehensive way.

To be specific, in the direction view, we evenly partition the space into a set of sectors $\mathcal{S} = \{s_k \, | \, k=0,1,...,n_s-1\}$, where $n_s$ denotes the number of partitioned sectors, which can be appropriately set for different datasets. Nodes in each direction-aware neighborhood $\mathcal{N}_{s_k}\!(v_i)$ of sector $s_k$ are grouped together, which we still call spatial group.
In the distance view, the space will be partitioned into $n_r$ distance rings $\mathcal{R} = \{r_k \, | \, k=0,1,...,n_r-1\}$, which is resulted from a predefined distance bucket (e.g., $<1km$, $1-2km$, and $>2km$).
Note that the central node $v_i$ itself does not belong to any sector or ring, we regard it as an additional group $\mathcal{N}_{s_{n_s}}\!(v_i) = \mathcal{N}_{r_{n_r}}\!(v_i) = \{v_i\}$.

\begin{figure}[t]
\centering
\includegraphics[width=0.95\columnwidth]{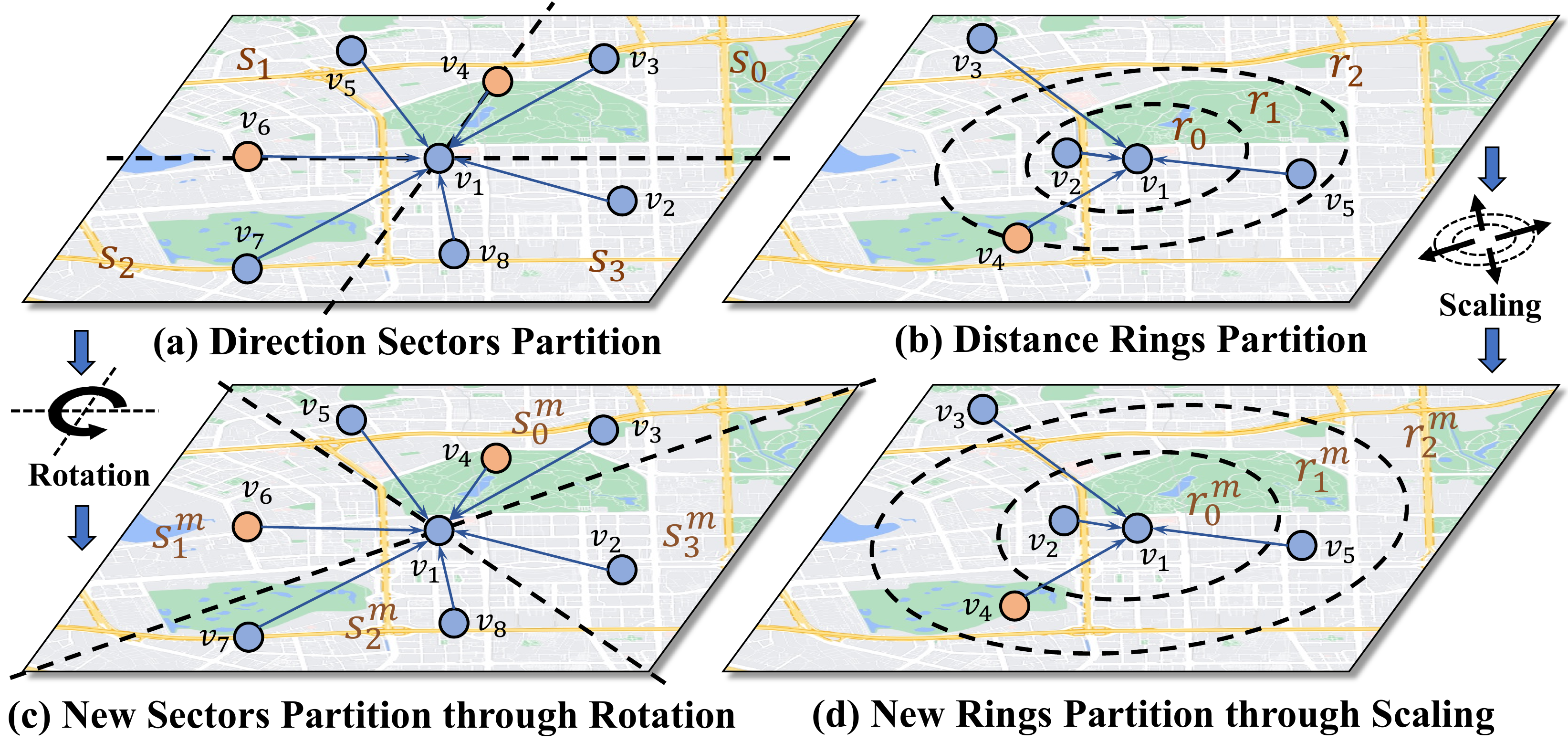}
 \vspace{-3mm}
\caption{Illustration of Rotation-scaling Partition.}
\label{fig_multi_head}
\vspace{-4mm}
\end{figure}

\vspace{-1mm}
\paragraph{Rotation-Scaling Multi-Head Partition.}
In view of the special case that a part of neighbors may locate at the boundary between two subspaces, we further propose a multi-head partition strategy to simultaneously perform multiple partitions at each view, where different heads can complement each other. 
For example, as shown in Figure \ref{fig_multi_head}(a), the orange node $v_4$ locates at the boundary between sectors $s_0$ and $s_1$, which indicates that the spatial relation of the neighborhood is still inadequately excavated by the single direction-based partition. A similar situation can be found in the partition of distance rings, such as the node $v_4$ in Figure \ref{fig_multi_head}(b).

To overcome this limitation, we extend our partition strategy by devising two operations, which are \textit{sector rotation} and \textit{ring scaling}, to achieve multiple space partitions for capturing the diverse spatial relations comprehensively.
Specifically, as illustrated in Figure \ref{fig_multi_head}(a) and (c), for the originally partitioned direction sectors, we turn the sector boundary a certain angle (e.g., 45 degrees) to derive another set of sectors, then the neighbors can be correspondingly reassigned in these new sectors and form a different set of direction-aware neighborhoods. Thus, we update the denotation of sectors as $\mathcal{S}^m \!=\! \{s^m_k \, | \, k=0,1,...,n_s\}$, and that of direction-aware neighborhoods as $\{\mathcal{N}_{s^m_k}(v_i) \, | \, k=0,1,...,n_s\}$, where $m=1,2,...,M_s$ denotes the $m$-th head partition among total $M_s$ heads. 
Similarly, from the distance view, we scale the boundary of the original distance rings to obtain the supplemental partition, which is shown in Figure \ref{fig_multi_head}(b) and (d). The denotations are updated as $\mathcal{R}^m \!=\! \{r^m_k \, | \, k=0,1,...,n_r\}$ and $\{\mathcal{N}_{r^m_k}(v_i) \, | \, k=0,1,...,n_r\}$ for rings and distance-aware neighborhoods, where $m=1,2,...,M_r$.
In this way, different heads of partitions model the spatial relation between neighbors and the central node complementarily, and thus we can avoid the improper grouping under a single partition.

\vspace{-1mm}
\subsubsection{Spatial-Aware Aggregation.}
\begin{figure*}[t]
\centering
\includegraphics[width=0.85\textwidth]{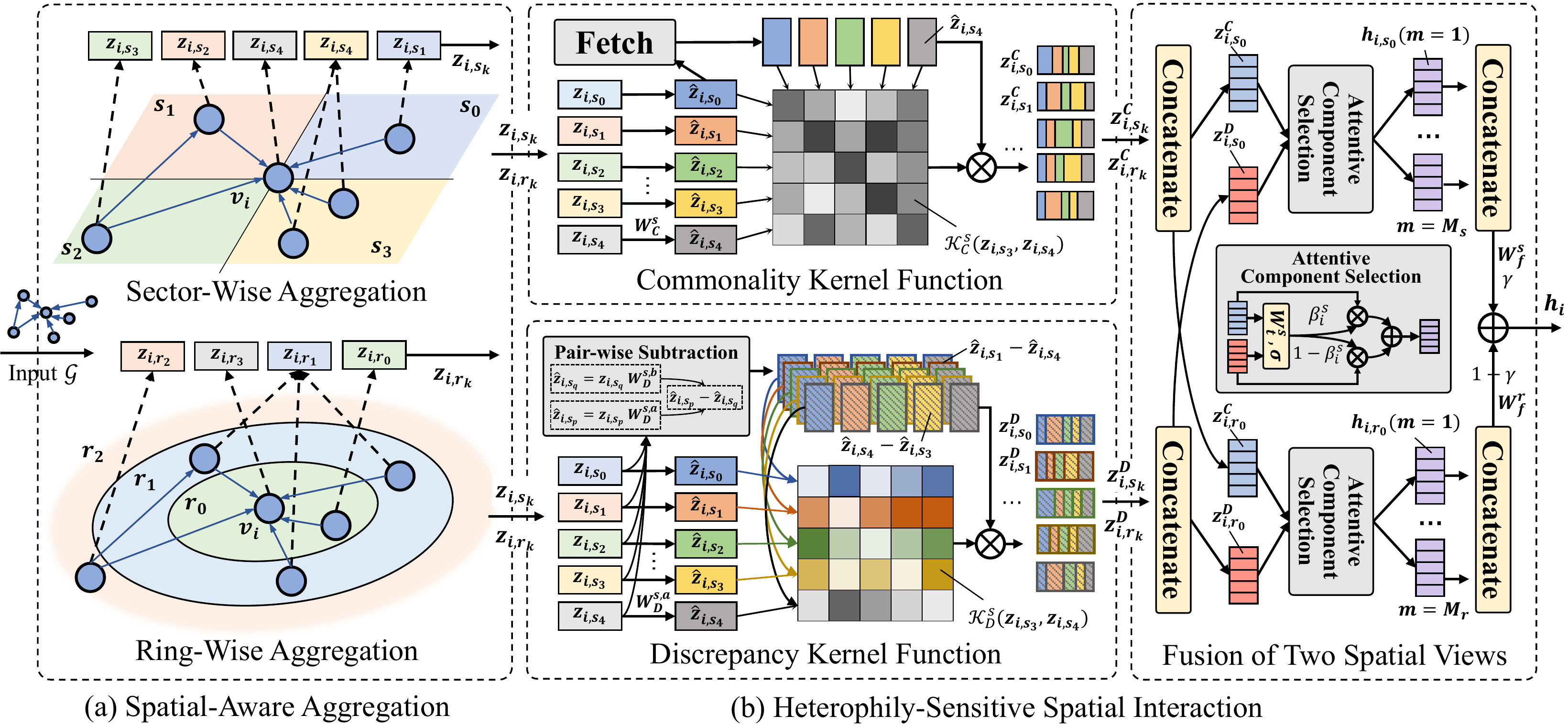}
 \vspace{-2mm}
\caption{The architecture of \mymodel. We only detailedly present the operation under sector partition in the illustration of two kernel functions and the attentive component selection. The same operation is also performed under ring partition.}
 \vspace{-4mm}
\label{fig_framework}
\end{figure*}

After the multi-head partition from two spatial perspectives, we collect messages from neighbors to the central node.
Rather than mixing the messages (e.g., through averaging) that most of the GNNs follow \cite{yang2021diverse}, our model performs a group-wise aggregation to handle different spatial groups with diverse heterophily. An illustration is shown in Figure \ref{fig_framework}(a).

Formally, taking the direction view as an example, with the set of direction-aware neighborhoods (spatial groups) under the $m$-head partition $\{\mathcal{N}_{s^m_k}(v_i) \, | \, k=0,1,...,n_s\}$, we use graph convolution \cite{kipf2016semi} to respectively aggregate the features of nodes in each neighborhood $\mathcal{N}_{s^m_k}(v_i)$ with the normalization by degree:
\begin{equation}
    \bm{z}_{i,s_k^m}(l+1) = \sum_{j\in \mathcal{N}_{s^m_k}(v_i)} (|\mathcal{N}(v_i)| \! \cdot \! |\mathcal{N}(v_j)|)^{-\frac{1}{2}} \, \bm{h}_{j}(l) \, \bm{W}_{s_k^m}(l),
\vspace{-1mm}
\end{equation}
where $\bm{z}_{i,s_k^m}(l)$ denotes the $l$-layer aggregated message from direction sector $s_k^m$, $\bm{h}_j(l)$ denotes the $l$-layer features of neighbor $v_j$ with $\bm{h}_j(0)=\bm{x}_j$, and $\bm{W}_{s_k^m}(l)$ is a trainable transformation extracting useful information from neighbors' features.

Similarly, at the distance view, we also perform the ring-wise aggregation \hide{with graph convolution }in each distance ring separately by:
\begin{equation}
    \bm{z}_{i,r_k^m}(l+1) = \sum_{j\in \mathcal{N}_{r^m_k}(v_i)} (|\mathcal{N}(v_i)| \! \cdot \! |\mathcal{N}(v_j)|)^{-\frac{1}{2}} \, \bm{h}_{j}(l) \, \bm{W}_{r_k^m}(l),
\vspace{-1mm}
\end{equation}
where $\bm{W}_{r_k^m}(l)$
% $\bm{W}_{r_k^m}^{(l)}$
is another feature transformation for neighbors at different distances. 
In this way, the aggregated messages can not only capture the structure information on the graph, but also discriminate their different spatial groups. It avoids losing different distributions of spatial heterophily on the urban graph.

\vspace{-1mm}
\subsection{Heterophily-Sensitive Spatial Interaction}
\label{heter_aware}
After the group-wise feature aggregation, \mymodel further captures the diverse spatial heterophily in different spatial groups on the urban graph.
In detail, we devise two learnable kernel functions to first respectively capture the commonality and discrepancy between the central node and each spatial group. Then, an attentive gate is jointly learned to adaptively determine the ratio of two components that should be propagated to the central node. Additionally, as indicated by the analysis of spatial tendency in Figure \ref{data_analysis}, the discrepancy of heterophily distributions varies along with the distance between two spatial groups, we further consider such characteristics by allowing the kernel functions to act between groups, which encourages the propagation of common and discrepant information among them. Since we view the central node as an additional group, this process can be regarded as an interaction among every two groups.
For simplicity, we omit the index $l$ and $m$ of layer and head in the following discussion.

\vspace{-1mm}
\paragraph{Commonality Kernel Function.}
Given the fact that different sectors / rings all belong to the neighborhood of the central node, they may share some common knowledge that can probably enhance the representation of each other. Thus, we first design a commonality kernel function to capture such information among them.

Formally, taking the direction view as an example, with the representation $\{\bm{z}_{i,s_k} \,|\, k=0,1,...,n_s\}$ of $n_s$ sectors centered by $v_i$, the kernel function $\mathcal{K}_C^s(\cdot, \cdot)$ that models the commonality degree between sector $s_p$ and $s_q$ (including $v_i$ itself) is defined as:
\begin{equation}
    \mathcal{K}_C^s(\bm{z}_{i,s_p}, \bm{z}_{i,s_q}) \,=\,\,<\!\hat{\bm{z}}_{i,s_p}, \hat{\bm{z}}_{i,s_q} \!\!>, \,\,\, \hat{\bm{z}}_{i, s_k} \!= \bm{z}_{i, s_k} \bm{W}_C^s,
\end{equation}
where $<\!\!\!\cdot, \cdot\!\!\!>$ denotes the inner product and $\bm{W}_C^s$ is the learnable matrix for common knowledge extraction. The larger output value suggests a higher similarity (i.e., commonality) between inputs.
Based on this measurement, we enhance the sector representation with the extracted useful information from other sectors:
\vspace{-0.5mm}
\begin{align}
    \bm{z}_{i, s_p}^C &= \sum_{q=0}^{n_s} \, \alpha_{pq}^{C,s} \, \bm{z}_{i, s_q} \bm{W}_C^s, \\
    \alpha_{pq}^{C,s} &= \frac{exp(\,\mathcal{K}_{C}^{s}(\bm{z}_{i,s_p}, \bm{z}_{i,s_q}))}
    {\,\sum_{k=0}^{n_s} exp(\mathcal{K}_{C}^{s}(\bm{z}_{i,s_p}, \bm{z}_{i,s_k}))},
\end{align}
where the coefficient $\alpha_{pq}^{C,s}$ is the level of commonality normalized by the softmax function. 
In the same way, we can also obtain the representation of each distance ring $\bm{z}_{i,r_p}^C$ with the enhancement of common knowledge from other rings using a similar kernel function $\mathcal{K}_C^r(\cdot, \cdot)$ parametrized by $\bm{W}_C^r$.

\vspace{-1mm}
\paragraph{Discrepancy Kernel Function.}
In addition to the common knowledge, modeling the difference information is critical on heterophilic urban graphs. Thus, we devise another kernel function to capture diverse dissimilarity between the central node and every group, as well as between any two groups. We introduce \hide{elaborate on }it from the direction view.
Specifically, taking the original representation $\{\bm{z}_{i,s_k} | k=0,1,...,n_s\}$ as inputs, the discrepancy kernel is defined as:
\begin{equation}
    \mathcal{K}_D^s(\bm{z}_{i,s_p}, \bm{z}_{i,s_q}) \,=\,\, <\! \bm{z}_{i, s_p} \bm{W}_D^{s,a}, ( \bm{z}_{i, s_p} \bm{W}_D^{s,a} - \bm{z}_{i, s_q} \bm{W}_D^{s,b})\!>,
\label{eq_discrepancy_kernel}
\end{equation}
where $\bm{W}_D^{s,a}$ and $\bm{W}_D^{s,b}$ denote two transformations that learn to extract the difference of sector $s_q$ compared to sector $s_p$.
According to \cite{hou2020measuring}, the kernel function $\mathcal{K}_D^s(\bm{z}_{i,s_p}, \bm{z}_{i,s_q}) $ can be regarded as calculating the dissimilarity degree between two inputs, which tends to be higher when $s_p$ and $s_q$ are more dissimilar.
Subsequently, our model facilitates each sector to be aware of the helpful difference information from others with this measurement of discrepancy:
\vspace{-1mm}
\begin{align}
    \bm{z}_{i,s_p}^D &= \sum_{q=0}^{n_s} \, \alpha_{pq}^{D,s} \, (\bm{z}_{i,s_p} \bm{W}_D^{s,a} \!- \bm{z}_{i,s_q} \bm{W}_D^{s,b}), \\
    \alpha_{pq}^{D,s} &= \frac{exp(\,\mathcal{K}_D^s(\bm{z}_{i,s_p}, \bm{z}_{i,s_q}))}
    {\,\sum_{k=0}^{n_s} exp(\mathcal{K}_D^s(\bm{z}_{i,s_p}, \bm{z}_{i,s_k}))}.
\end{align}
Similarly, we utilize an analogous kernel function, which is denoted as $\mathcal{K}_D^r(\cdot, \!\cdot)$ with parameters $\bm{W}_D^{r,a}$ and $\bm{W}_D^{r,b}$ to capture such discrepancy at the distance view and derive the ring representation $ \bm{z}_{i,r_p}^D$ aware of the diverse distributions of spatial heterophily.

\vspace{-1mm}
\paragraph{Attentive Component Selection.}
With these two kernel functions, we can exploit both the common knowledge and diverse discrepancy information between the central node and neighbors in different groups. However, different nodes may possess varying levels of spatial heterophily in various applications. Thus, \mymodel learns to derive a gate that adaptively determines the ratio of common and difference information \hide{the central node needs, }in an end-to-end manner.

Specifically, for each central node $v_i$, we concatenate both the commonality and discrepancy components of all sectors, to derive a scalar with a transformation:
\begin{equation}
    \beta_i^{s} = \sigma(\parallel_{j\in \{C, D\}}\parallel_{k=0}^{n_s}\bm{z}_{i,s_k}^j \bm{W}_t^s) ,
\end{equation}
where $\bm{W}_t^s$ denotes the trainable transformation mapping the input to a scalar and $\sigma$ denotes the \textit{Sigmoid} function restricting the output value into $(0,1)$. 
Then, $\beta_i^s$ serves as a gate controlling the ratio of the commonality and discrepancy components in the final representation of each sector:
\begin{equation}
    \bm{\tilde{z}}_{i,s_k} = \beta_i^{s} \cdot \bm{z}_{i,s_k}^C + (1 - \beta_i^{s}) \cdot \bm{z}_{i,s_k}^D.
\end{equation}
In the same way, we learn to derive the gate $\beta_{i}^r$ that determines the ratio in each ring's final representation $\bm{\tilde{z}}_{i,r_k}$.

After the propagation among spatial groups (including the central node), different groups can contain diverse discrepancy information in the neighborhood, which is vital to the heterophilic urban graph. We then integrate this group-wise representation by concatenation (instead of summation, to avoid the mixing of diverse distributions of spatial heterophily) to obtain the global representation of two views. Since we adopt the multi-head partition strategy, a following concatenation is used to combine different heads at each view. The above two processes can be jointly expressed as:
\begin{equation}
    \bm{h}_{i,s} = \, \parallel_{m=1}^{M_s} \, (\parallel_{k=0}^{n_s} \, \bm{\tilde{z}}_{i,s_k^m}), \,\,\,
    \bm{h}_{i,r} = \, \parallel_{m=1}^{M_r} \, (\parallel_{k=0}^{n_r} \, \bm{\tilde{z}}_{i,r_k^m}).
\end{equation}

\paragraph{Fusion of Two Spatial Views.}
Finally, we fuse two spatial views with a learnable weighted summation to update the central node's representation as follows:
\begin{equation}
    \bm{h}_i = \gamma \, \bm{h}_{i,s} \bm{W}_f^s + (1-\gamma) \, \bm{h}_{i,r} \bm{W}_f^r,
\end{equation}
where $\bm{W}_f^s$ and $\bm{W}_f^r$ are two weight matrices transforming the representation vectors of two views into the same space, and $\gamma$ is a trainable trade-off parameter activated by \textit{Sigmoid} function, which learns to assign different importance to the direction and distance view according to the target task. 

\subsection{Prediction and Optimization}
Consistent with general GNNs, we use the $L$-layer output activated by \textit{ReLU} function $\hat{\bm{h}_i} \!= \!\sigma(\bm{h}_i^{(L)})$ as the node representation to make a prediction in different downstream tasks, and optimize the model by the appropriate loss function: $\mathcal{L}(\,LR(\hat{\bm{h}_i})\,,\, y_i\,)$.
In node regression tasks, $LR(\cdot)$ is the linear regressor, $y_i\!\in \!\mathbb{R}$ denotes the ground truth of labeled nodes, and $\mathcal{L}$ can be L2 loss.
While for node classification tasks, $LR(\cdot)$ performs the logistic regression, $y_i\! \in \!\{0,1\}^{\mathcal{C}}$ is a one-hot label vector of $\mathcal{C}$ classes, and $\mathcal{L}$ can be cross entropy loss.

\section{Experiments}
\label{section_exp}
In this section, we conduct extensive experiments on real-world datasets in three different tasks upon two types of urban graphs to evaluate the effectiveness of our model.
The code of \mymodel is available at 
\url{https://github.com/PaddlePaddle/PaddleSpatial/tree/main/research/SHGNN}.

% \vspace{-1.5mm}
\subsection{Experiment Settings}
\subsubsection{Tasks and Data Description.}
\label{exp_task_description}
We first briefly introduce the three datasets corresponding to three different tasks, and details of how to build each dataset are described in Appendix \ref{apdx_dataset}. Table \ref{exp-dataset} summarizes the statistical information of the three datasets. 

\renewcommand\arraystretch{0.9}
\begin{table}[t]
    \small
    \caption{Statistics of three real-world datasets.}
    \vspace{-4mm}
    \label{exp-dataset}
    \centering
    \resizebox{0.46\textwidth}{!}{
    % \begin{tabular}{c|c|c|c|c|c|c|c|c}
    \begin{tabular}{ccccccc}
	\toprule
	\textbf{Task} & \textbf{Dataset} & \# \textbf{Nodes} & \# \textbf{Edges} & \# \textbf{Labeled} & $\bm{\lambda_d^s}$ & $\bm{\lambda_d^r}$ \\
	% \cline{1-9}
	% \rule{0pt}{10pt}
        \midrule
	CAP & Shenzhen & 82,510 & 9,486,879 & 9,121 & 0.89 & 0.78 \\
	% \cline{1-9}
	% \rule{0pt}{10pt}
	CP & Manhattan & 180 & 3,780 & 180 & 0.98 & 0.98 \\
	% \cline{1-9}
	% \rule{0pt}{10pt}
	DRSD & Los Angeles & 253,985 & 1,365,289 & 15,274 & 0.23 & 0.14 \\
	\bottomrule
    \end{tabular}
    }
    \vspace{-4mm}
\end{table}

% \renewcommand\arraystretch{0.9}
% \begin{table}[t]
%     \small
%     \caption{Statistics of three real-world datasets.}
%     \vspace{-4mm}
%     \label{exp-dataset}
%     \centering
%     \resizebox{0.48\textwidth}{!}{
%     % \begin{tabular}{c|c|c|c|c|c|c|c|c}
%     \begin{tabular}{ccccccccc}
% 	\toprule
% 	\textbf{Task} & \textbf{Dataset} & \# \textbf{Nodes} & \# \textbf{Edges} & \# \textbf{Labeled} & $\bm{\lambda_d^s}$ & $\bm{\lambda_d^r}$ & $\bm{\lambda_t^s}$ & $\bm{\lambda_t^r}$ \\
% 	% \cline{1-9}
% 	% \rule{0pt}{10pt}
%         \midrule
% 	CAP & Shenzhen & 82,510 & 9,486,879 & 9,121 & 0.89 & 0.78 & 0.85 & 0.92\\
% 	% \cline{1-9}
% 	% \rule{0pt}{10pt}
% 	CP & Manhattan & 180 & 3,780 & 180 & 0.98 & 0.98 & 0.72 & 0.93\\
% 	% \cline{1-9}
% 	% \rule{0pt}{10pt}
% 	DRSD & Los Angeles & 253,985 & 1,365,289 & 15,274 & 0.23 & 0.14 & 0.53 & 0.52\\
% 	\bottomrule
%     \end{tabular}
%     }
%     \vspace{-5mm}
% \end{table}
% \begin{table}[t]
%     \small
%     \caption{Statistics of three real-world datasets.}
%     \vspace{-3mm}
%     \label{exp-dataset}
%     \centering
%     \resizebox{0.4\textwidth}{!}{
%     \begin{tabular}{c|c|c|c|c}
% 	\toprule
% 	Task & Dataset & \# Nodes & \# Edges & \# Labeled Nodes \\
% 	\cline{1-5}
% 	\rule{0pt}{10pt}
% 	CAP & Shenzhen & 82,510 & 9,486,879 & 9,121\\
% 	\cline{1-5}
% 	\rule{0pt}{10pt}
% 	CP & Manhattan & 180 & 3,780 & 180\\
% 	\cline{1-5}
% 	\rule{0pt}{10pt}
% 	DRSD & Los Angeles & 253,985 & 1,365,289 & 15,274\\
% 	\bottomrule
%     \end{tabular}
%     }
%     \vspace{-4mm}
% \end{table}
% num_nodes=82510, num_edges=9486879
% num_nodes=901800, num_edges=2233510

\textbf{Commercial Activeness Prediction (CAP)}
is a node regression task on a mobility graph. Similar to \cite{xi2022beyond}, we use the number of comments to POIs in each region as the indicator of regional commercial activeness.
To form the dataset of this task, we collect the following urban data in \textbf{\textit{Shenzhen}} city in China from Baidu Maps, including POI data and satellite images in September 2019 to construct region features, the daily human flow data from July 2019 to September 2019 for building the edge set of the urban graph, and the number of regional POI comments from June 2019 to April 2020 which is regarded as the ground truth. 

\renewcommand\arraystretch{0.9}
\begin{table*}[t]
    \caption{Performance comparison. `TASK-S' refers to task-specific baseline for CAP (KnowCL), CP (NNCCRF), and DRSD (RFN). Symbol $\bm{\ast}$, $\blacktriangle$ and $\vartriangle$ indicate that \mymodel achieves significant improvements with $p<0.001$, $p<0.01$ and $p<0.05$, respectively.}
    \small
    \vspace{-4mm}
    \centering
    \begin{tabular}{p{1.1cm}<{\centering} | p{1.6cm}<{\centering} p{1.6cm}<{\centering} p{1.6cm}<{\centering} | p{1.6cm}<{\centering} p{1.6cm}<{\centering} p{1.65cm}<{\centering} | p{1.9cm}<{\centering} p{1.9cm}<{\centering}}
    \toprule
     & \multicolumn{3}{c|}{Commercial Activeness Prediction} & \multicolumn{3}{c|}{Crime Prediction} & \multicolumn{2}{c}{Dangerous Road Section Detection} \\
    \cline{1-9}
    \rule{0pt}{10pt}
    Methods & RMSE $\downarrow$ & MAE $\downarrow$ & R$^2 \uparrow$ & RMSE $\downarrow$ & MAE $\downarrow$ & R$^2 \uparrow$ & AUC $\uparrow$ & F1-score $\uparrow$\\
    \midrule
    GCN & 8.354 ± 0.020\textbf{*} & 5.018 ± 0.028\textbf{*} & 0.388 ± 0.003\textbf{*} & 156.1 ± 1.312\textbf{*} & 114.5 ± 1.005\textbf{*} & 0.082 ± 0.015\textbf{*} & 0.634 ± 0.002\textbf{*} & 0.229 ± 0.003\textbf{*}\\
    GAT & 8.952 ± 0.153\textbf{*} & 5.134 ± 0.099\textbf{*} & 0.298 ± 0.023\textbf{*} & 166.8 ± 5.750\textbf{*} & 127.1 ± 4.359\textbf{*} & -0.040 ± 0.071\textbf{*} & 0.613 ± 0.006\textbf{*} & 0.213 ± 0.008$^\blacktriangle$\\
    % \midrule
    Mixhop & 8.168 ± 0.038\textbf{*} & 4.981 ± 0.010\textbf{*} & 0.415 ± 0.005\textbf{*} & 147.6 ± 0.576\textbf{*} & 109.0 ± 0.398\textbf{*} & 0.179 ± 0.006\textbf{*} & 0.636 ± 0.003\textbf{*} & 0.229 ± 0.005\textbf{*}\\
    FAGCN & 8.327 ± 0.018\textbf{*} & 5.063 ± 0.031\textbf{*} & 0.392 ± 0.002\textbf{*} & 150.9 ± 3.618\textbf{*} & 108.1 ± 2.179$^\blacktriangle$ & 0.142 ± 0.041\textbf{*} & 0.632 ± 0.004\textbf{*} & 0.228 ± 0.009$^\blacktriangle$\\
    NLGCN & 8.495 ± 0.108\textbf{*} & 4.883 ± 0.053\textbf{*} & 0.367 ± 0.016\textbf{*} & 151.7 ± 4.496\textbf{*} & 110.2 ± 3.557$^\blacktriangle$ & 0.133 ± 0.050\textbf{*} & 0.626 ± 0.003\textbf{*} & 0.224 ± 0.004\textbf{*}\\
    GBKGNN & 8.490 ± 0.017\textbf{*} & 5.004 ± 0.074$^\blacktriangle$ & 0.368 ± 0.002\textbf{*} & 133.6 ± 4.095$^\blacktriangle$ & 98.28 ± 3.792 & 0.327 ± 0.041$^\blacktriangle$ & 0.626 ± 0.011\textbf{*} & 0.225 ± 0.013$^\blacktriangle$\\
    GPRGNN & 8.139 ± 0.016$^\blacktriangle$ & 4.975 ± 0.025\textbf{*} & 0.419 ± 0.002$^\blacktriangle$ & 152.4 ± 0.338\textbf{*} & 107.0 ± 0.319\textbf{*} & 0.126 ± 0.003\textbf{*} & 0.635 ± 0.003\textbf{*} & 0.226 ± 0.004\textbf{*}\\
    % \midrule
    PRIM & 8.928 ± 0.144\textbf{*} & 5.186 ± 0.086\textbf{*} & 0.301 ± 0.022\textbf{*} & 151.7 ± 6.243\textbf{*} & 118.6 ± 4.320\textbf{*} & 0.131 ± 0.070\textbf{*} & 0.615 ± 0.008\textbf{*} & 0.210 ± 0.005\textbf{*}\\
    SAGNN & 8.843 ± 0.068\textbf{*} & 5.305 ± 0.057\textbf{*} & 0.315 ± 0.010\textbf{*} & 142.2 ± 2.451\textbf{*} & 108.6 ± 1.360\textbf{*} & 0.238 ± 0.026\textbf{*} & 0.630 ± 0.006\textbf{*} & 0.216 ± 0.008\textbf{*}\\
    % \midrule
    TASK-S & 8.360 ± 0.068\textbf{*} & 4.875 ± 0.028\textbf{*} & 0.387 ± 0.010\textbf{*} & 138.8 ± 2.619\textbf{*} & 100.4 ± 0.945$^\blacktriangle$ & 0.274 ± 0.027\textbf{*} & 0.638 ± 0.012$^\blacktriangle$ & 0.240 ± 0.008$^\blacktriangle$\\
    % \midrule
    \mymodel & \textbf{7.605 ± 0.131} & \textbf{4.485 ± 0.056} & \textbf{0.493 ± 0.017} & \textbf{118.7 ± 4.135} & \textbf{91.16 ± 2.754} & \textbf{0.469 ± 0.036} & \textbf{0.684 ± 0.004} & \textbf{0.263 ± 0.003}\\
    \bottomrule
\end{tabular}
\vspace{-5.5mm}
\label{table-exp-main}
\end{table*}

\begin{figure*}[t]
\centering
    \subfigure{
    \includegraphics[width=0.345\columnwidth]{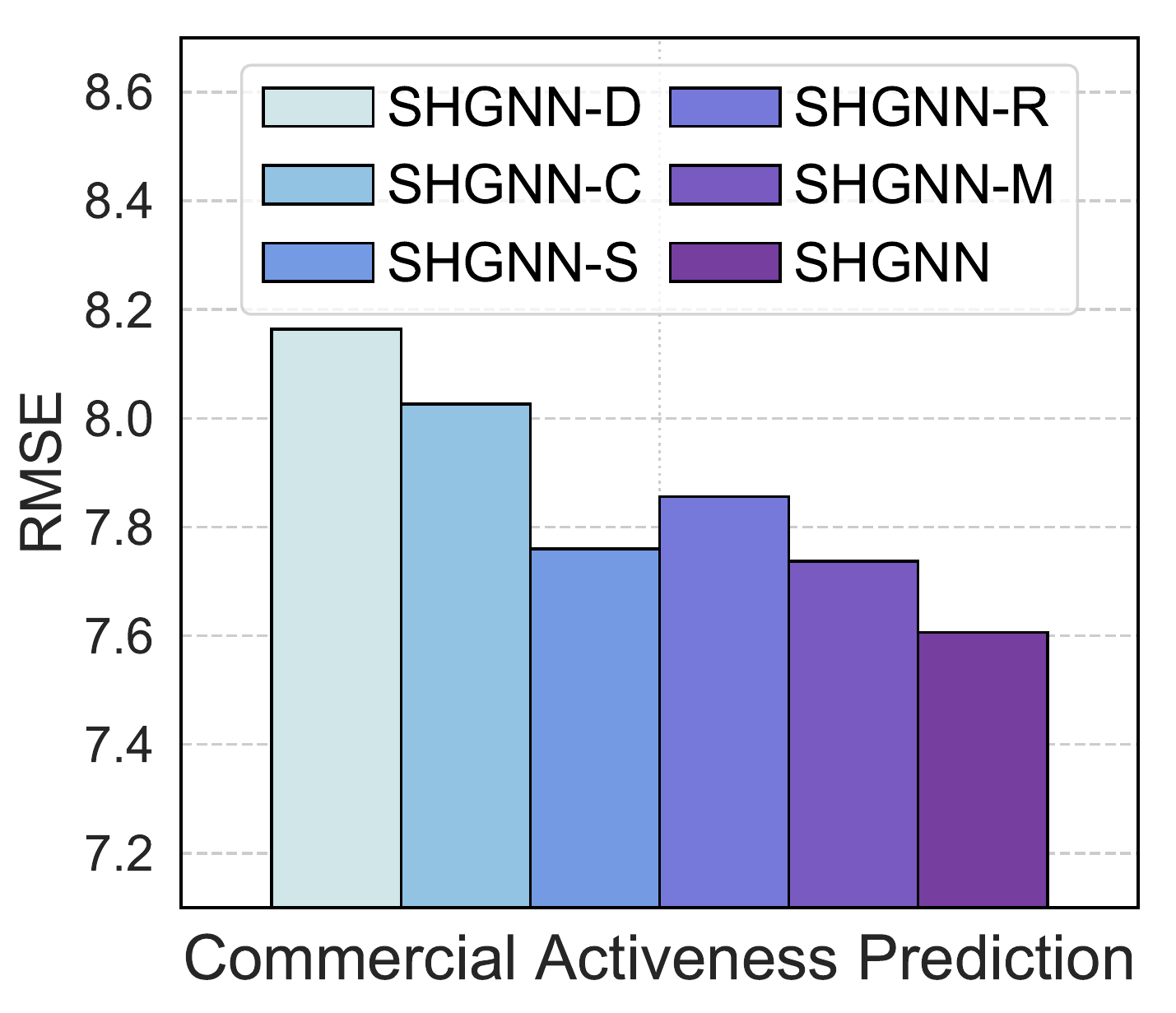}}
    \hspace{-2.5mm}
    \subfigure{
    \includegraphics[width=0.345\columnwidth]{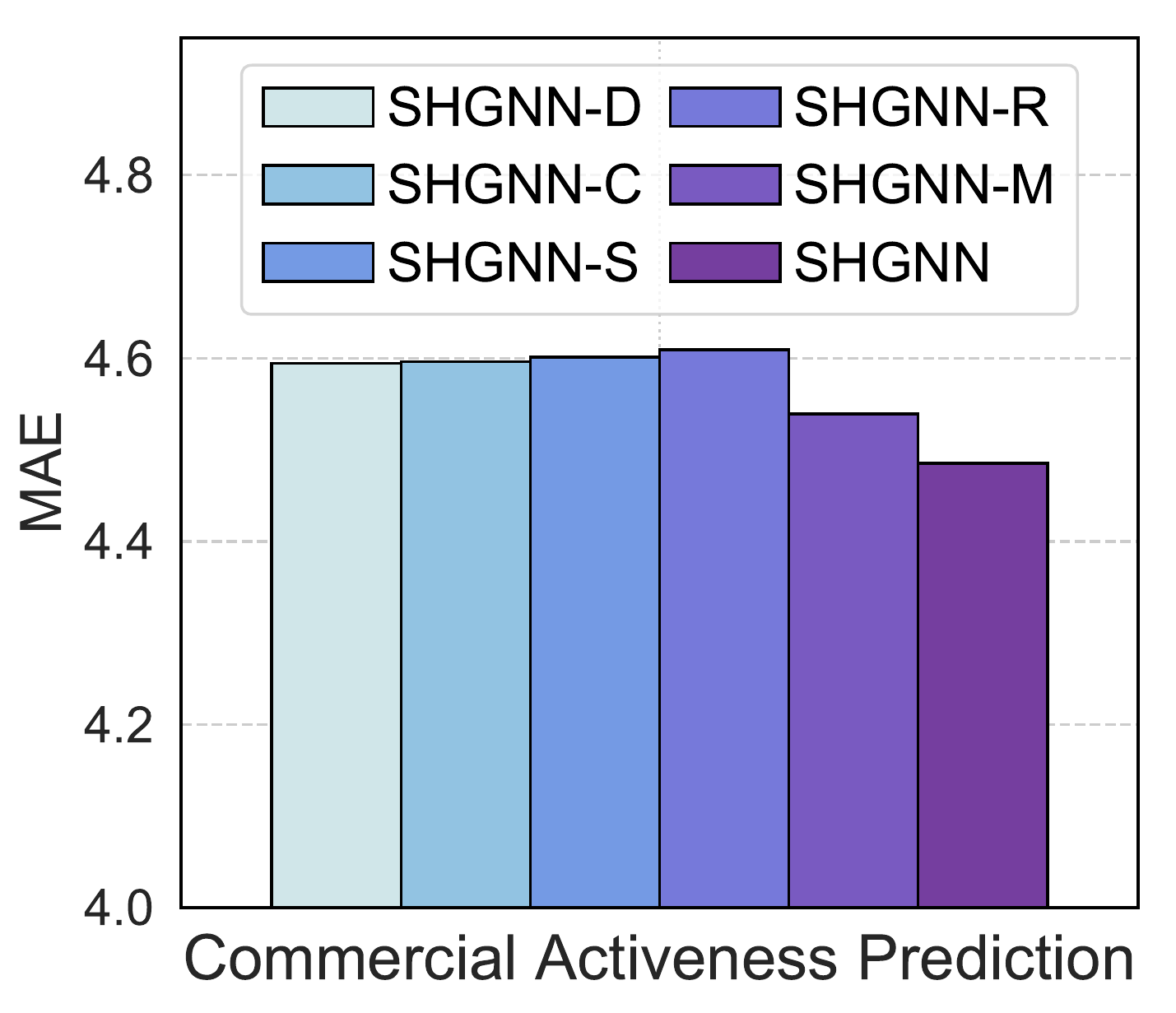}}
    \hspace{-2.5mm}
    \subfigure{
    \includegraphics[width=0.35\columnwidth]{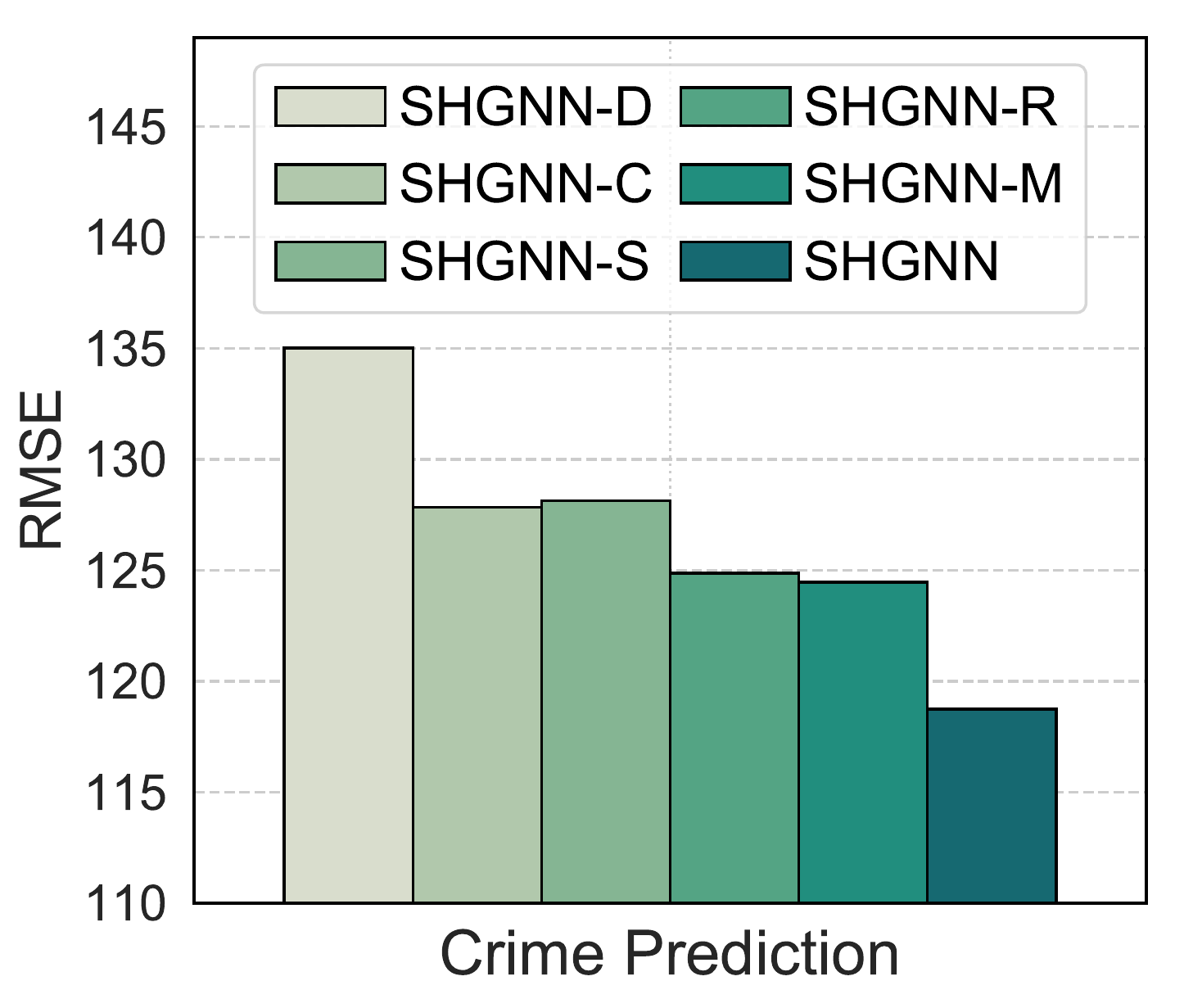}}
    \hspace{-2.5mm}
    \subfigure{
    \includegraphics[width=0.35\columnwidth]{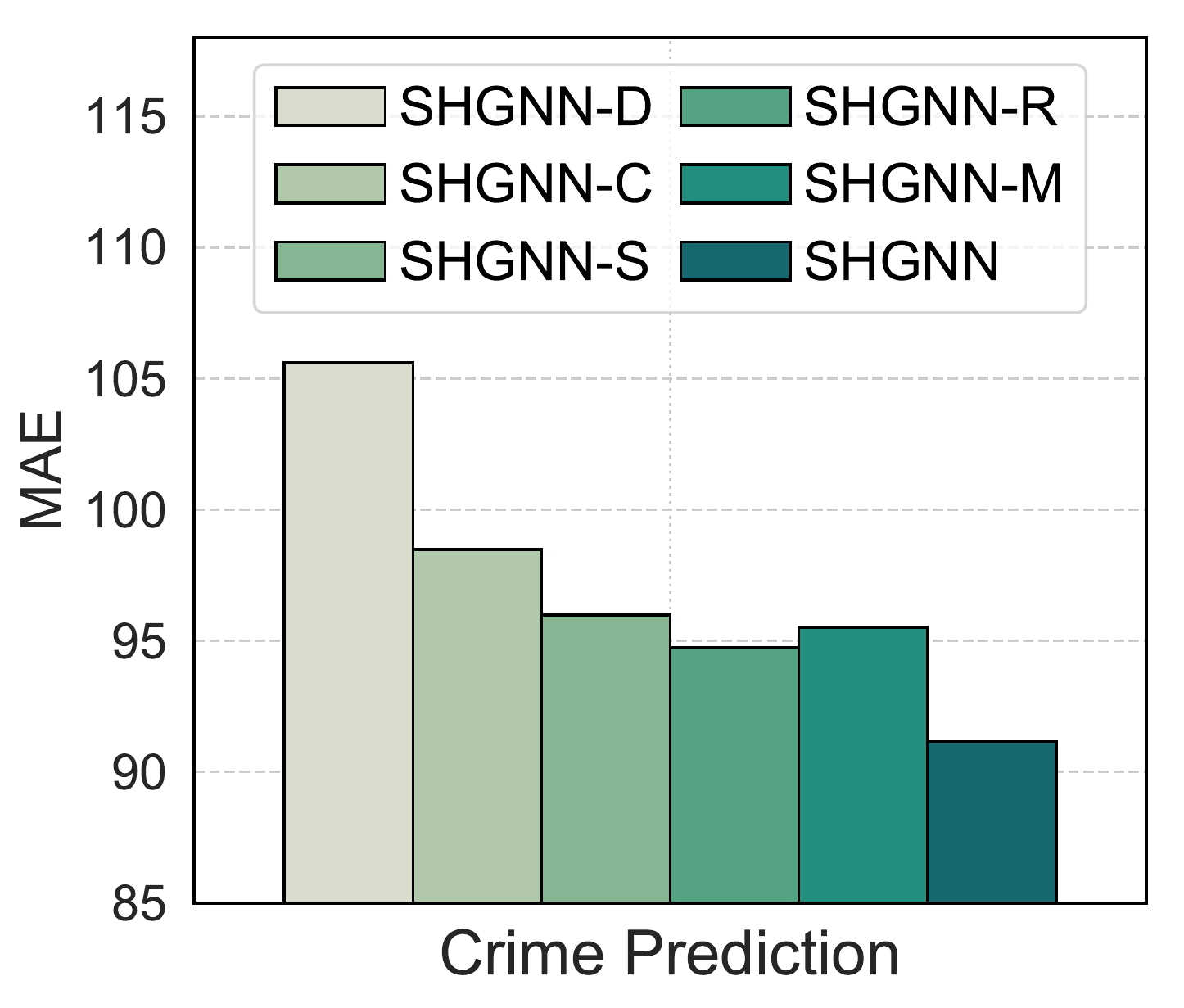}}
    \hspace{-2.5mm}
    \subfigure{
    \includegraphics[width=0.35\columnwidth]{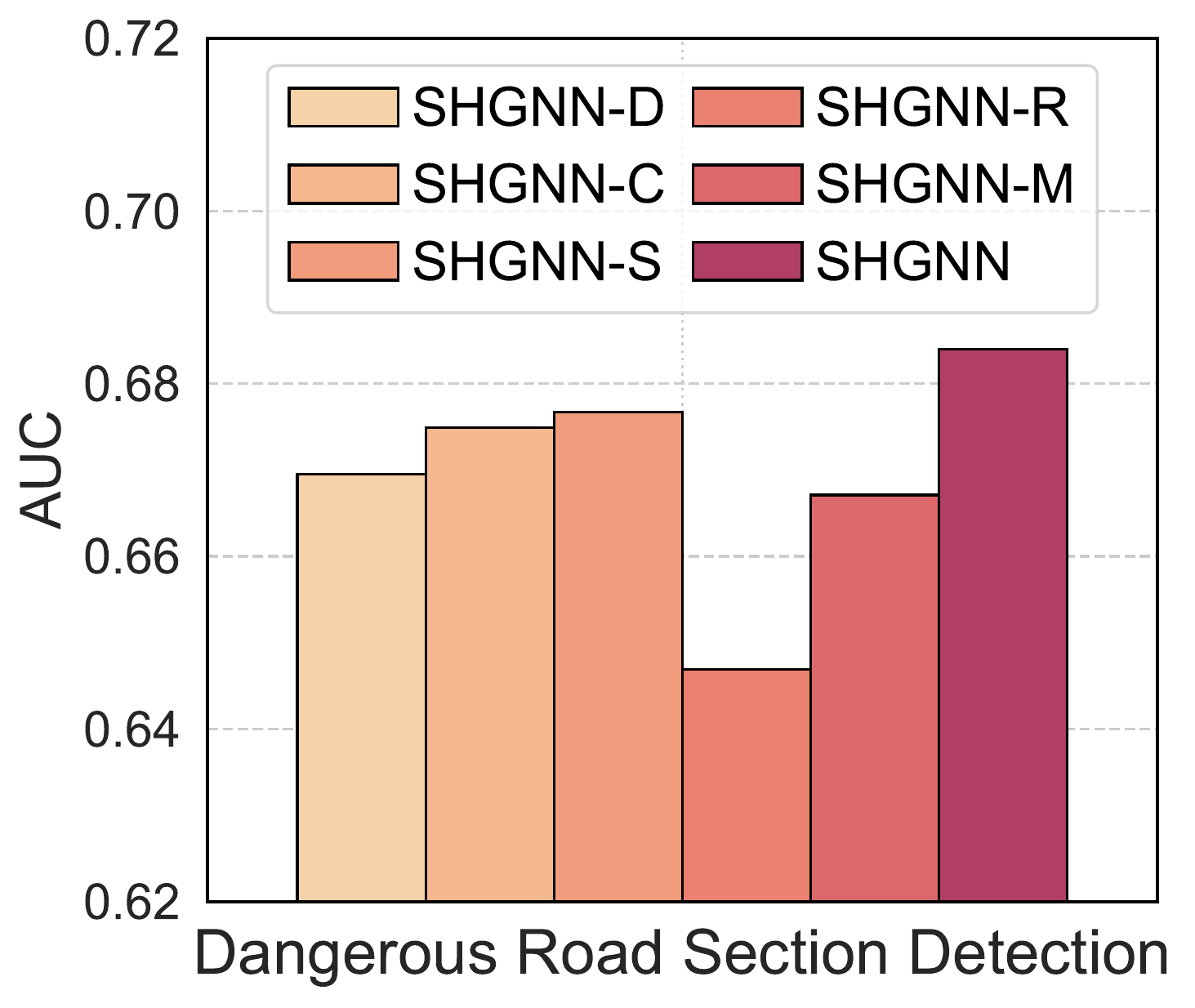}}
    \hspace{-2.5mm}
    \subfigure{
    \includegraphics[width=0.35\columnwidth]{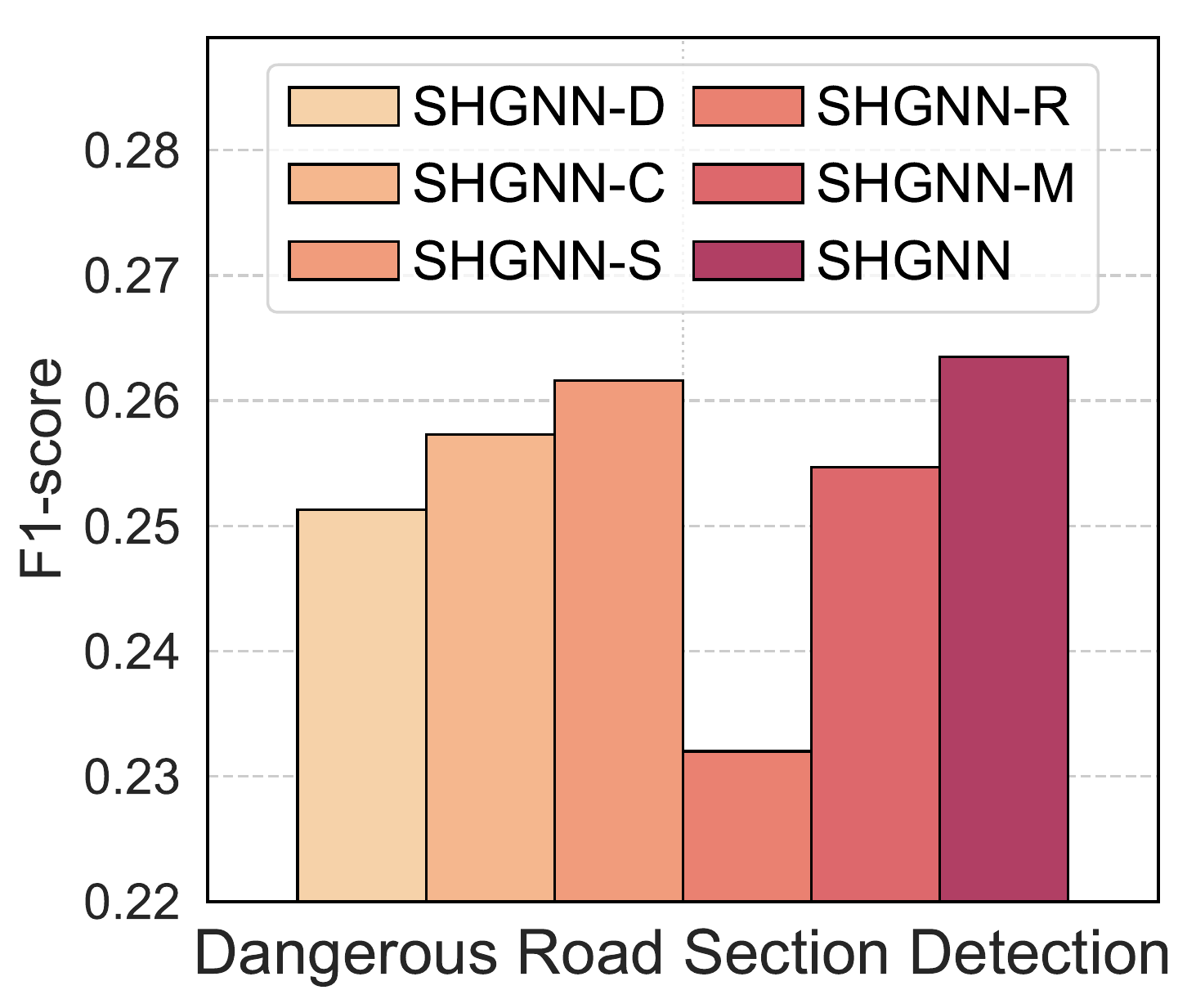}}
   \vspace{-6mm}
  \caption{Performance comparison between \mymodel and its variants on three tasks.}
\vspace{-5.5mm}
\label{fig_ablation}
\end{figure*}

\textbf{Crime Prediction (CP)}
is also a node regression task on the mobility graph. We collect the real-world dataset of \textbf{\textit{New York City}} from NYC open data website\footnote{opendata.cityofnewyork.us} for this task. The dataset contains 180 regions in Manhattan, with the POI data and crime number in each region, as well as taxi trips between regions \cite{zhang2021multi,wu2022multi}. We construct node features from POIs. Taxi trips are used to build the urban graph, where we only keep the 20 most important edges for each region w.r.t. the number of trip records.

\textbf{Dangerous Road Section Detection (DRSD)}
is a node classification task performed on the road network. In this work, the dangerous section is defined as the road section with a high incidence of traffic accidents. We build a real-world dataset in \textbf{\textit{Los Angeles}} based on the road network data from OSMnx Street Networks in Harvard Dataverse\footnote{https://dataverse.harvard.edu/dataverse/osmnx-street-networks} 
and the traffic accident records in December 2021 from Kaggle dataset website\footnote{https://www.kaggle.com/datasets/sobhanmoosavi/us-accidents}.
First, we make statistics of the number of accident records on each road section. Then, the sections containing more than 3 accident records in a month will be considered as dangerous road sections in our experiments.

To select the best hyper-parameters for all the comparing methods, we randomly split each dataset into three parts with $60\%$ for training, $20\%$ for validation and $20\%$ for test. 

\vspace{-1mm}
\subsubsection{Baselines.}
We compare \mymodel with a variety of state-of-the-art GNN models, including two classical message passing neural networks (\textbf{GCN} \cite{kipf2016semi} and \textbf{GAT} \cite{velivckovic2018graph}), five representative methods for heterophilic graphs (\textbf{Mixhop} \cite{abu2019mixhop}, \textbf{FAGCN} \cite{bo2021beyond}, \textbf{NLGCN} \cite{liu2021non}, \textbf{GPRGNN} \cite{chien2021adaptive} and \textbf{GBKGNN} \cite{du2022gbk}), two spatial GNN models
(\textbf{SAGNN} \cite{li2020competitive} and \textbf{PRIM} \cite{chen2021points}),
as well as three task-specific baselines (\textbf{KnowCL} \cite{liu2023knowledge} for CAP, \textbf{NNCCRF} \cite{yi2019neural} for CP and \textbf{RFN} \cite{jepsen2020relational} for DRSD).
Detailed descriptions are introduced in Appendix \ref{baseline_describe}.

\vspace{-1mm}
\subsubsection{Evaluation Metrics.}
For the two regression tasks, we evaluate all methods with Root Mean Square Error (RMSE), Mean Absolute Error (MAE) and the coefficient of determination (R$^2$). For the node classification task, we use Area Under Curve (AUC) and F1-score.

\vspace{-1mm}
\subsection{Performance Evaluation}
\subsubsection{Overall Comparison.}
The performance comparison of our \mymodel and baselines is presented in Table \ref{table-exp-main}, in which the mean and standard deviation of all metrics are obtained through five random runs.
As we can see, \mymodel consistently achieves the best performance in three tasks on two kinds of urban graphs, with $6.6\%$ and $11.2\%$ reductions of RMSE in commercial activeness prediction (CAP) and crime prediction (CP), as well as $7.2\%$ improvements of AUC in dangerous road section detection (DRSD) over the most competitive baseline of each task. We also conduct a pairwise t-test between \mymodel and each baseline to demonstrate that our model outperforms all of them significantly.
Note that although the spatial diversity of heterophily on the road network in DRSD task is not so strong ($\lambda_d^s=0.23$ and $\lambda_d^r=0.14$) as another two urban graphs, our model can still improve the accuracy in a large margin. It indicates the effectiveness of \mymodel can be general but not just limited to urban graphs with strong spatial heterophily.

Specifically, the ordinary GNN models (GCN and GAT) generally have the worst overall performance.
By contrast, approaches designed to deal with graph heterophily (Mixhop, FAGCN, NLGCN, GBKGNN and GPRGNN) evidently perform better. 
It indicates the inappropriateness of simply treating an urban graph as a general homophilic graph. However, as a specially designed model to handle spatial heterophily, our \mymodel remarkably outperforms these general heterophilic GNNs.
The spatial GNN methods (SAGNN and PRIM) perform better than GCN and GAT sometimes, but perform worse than the methods for heterophilic graphs in many cases. 
And for task-specific baselines, it can also be found that some heterophilic GNNs are level pegging with them (such as Mixhop and GPRGNN vs. KnowCL in CAP task, GBKGNN vs. NNCCRF in CP task and Mixhop vs. RFN in DRSD task). 
These results also demonstrate the importance to take the spatial heterophily into consideration when using GNNs over an urban graph.
To sum up, our \mymodel is much more effective in considering and alleviating the spatial heterophily on urban graphs in all the tasks.

\vspace{-1.5mm}
\subsubsection{Ablation Study.}
To verify the effectiveness of each design in our model, we further compare \mymodel with its five variants:
\begin{itemize}[leftmargin=*, topsep=2pt]
    \item \textbf{\mymodel-S} removes the direction section partition, which only models spatial heterophily from the distance view.
    \item \textbf{\mymodel-R} removes the distance ring partition,  which only models spatial heterophily from the direction view.
    \item \textbf{\mymodel-M} removes the multi-head partition strategy.
    \item \textbf{\mymodel-C} removes the commonality kernel function, without sharing common knowledge among spatial groups.
    \item \textbf{\mymodel-D} removes the discrepancy kernel function. It cannot capture the difference among spatial groups.
\end{itemize}
As shown in Figure \ref{fig_ablation}, \mymodel outperforms all the variants, proving the significance of our designs in tackling spatial heterophily. Specifically, the performances get worse if we remove either the partition of direction sections or distance rings (\mymodel-S and \mymodel-R), which indicates the necessity of considering the spatial heterophily from both views.
Besides, the rotation-scaling multi-head partition strategy can evidently help to model diverse spatial relations (\mymodel-M).
In addition, the performance degrades if the commonality kernel function is not used, suggesting the effectiveness of sharing knowledge among neighbors. More importantly, removing the discrepancy kernel function results in a notable performance decline, which verifies the importance of further capturing and exploiting the difference information on heterophilic urban graphs.

\begin{figure}[t]
\centering
\vspace{-1.5mm}
\subfigure{
    \includegraphics[width=0.45\columnwidth]{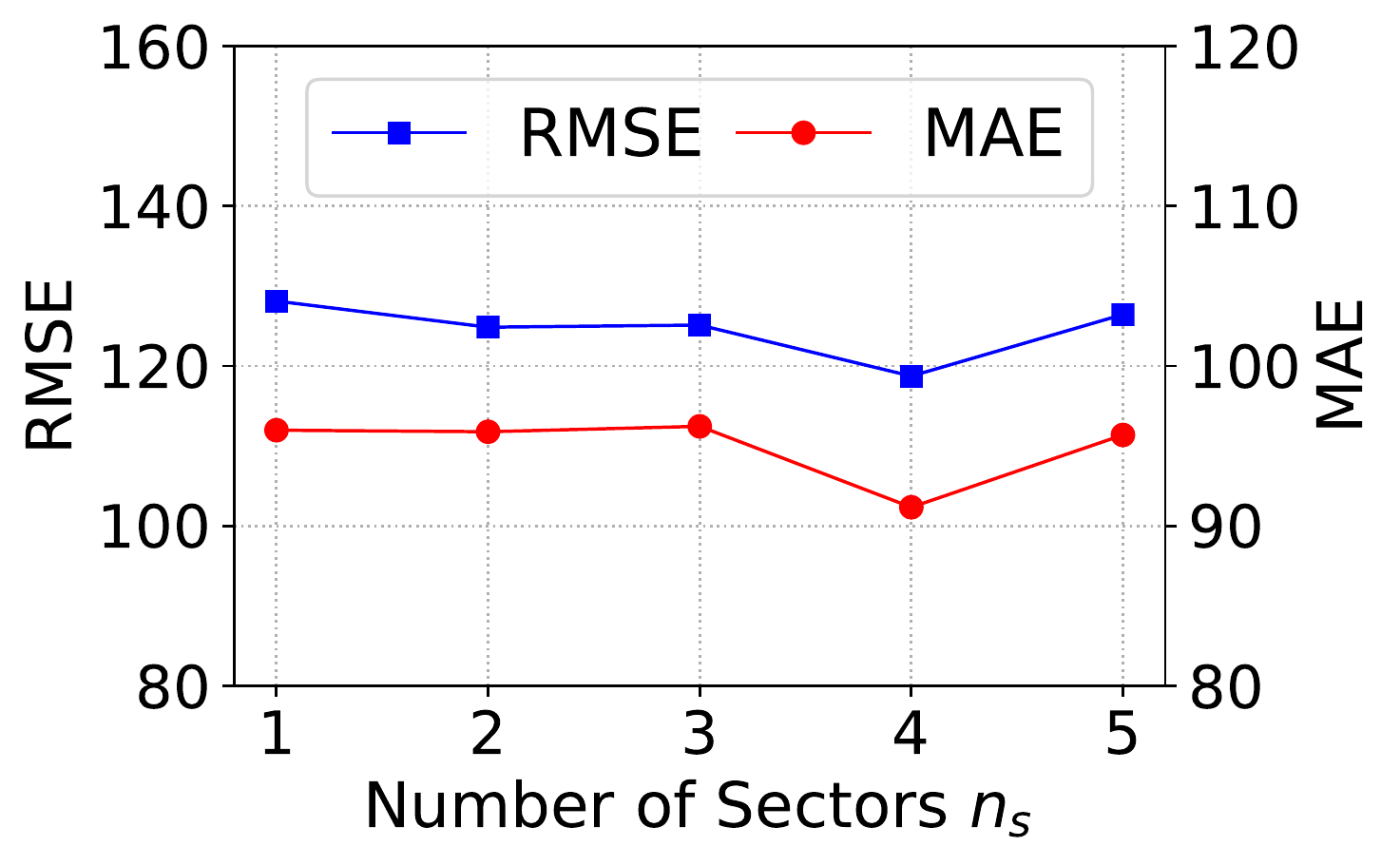}}
\subfigure{
    \includegraphics[width=0.45\columnwidth]{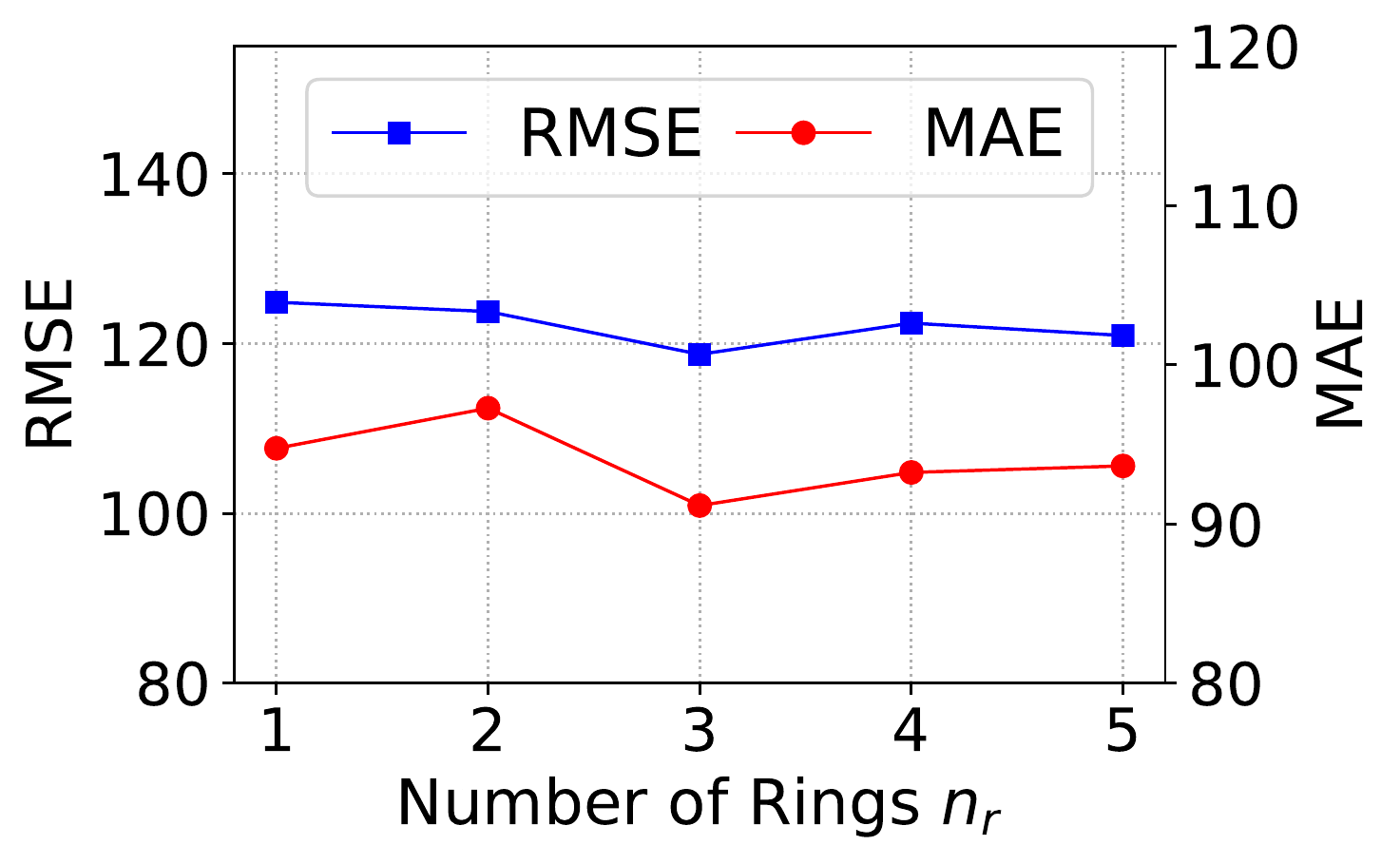}} \\
    \vspace{-4.5mm}
\subfigure{
    \includegraphics[width=0.45\columnwidth]{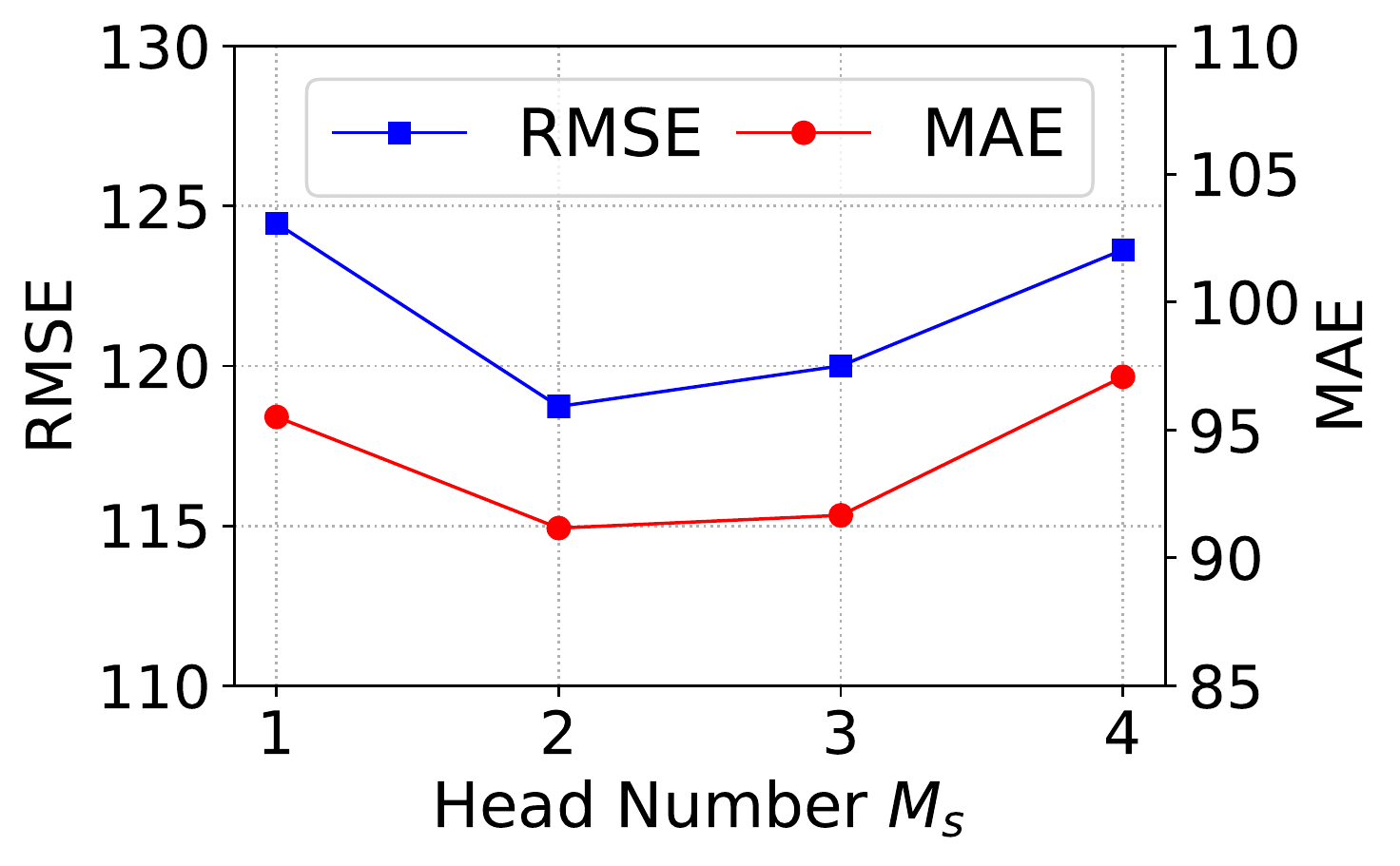}}
\subfigure{
    \includegraphics[width=0.45\columnwidth]{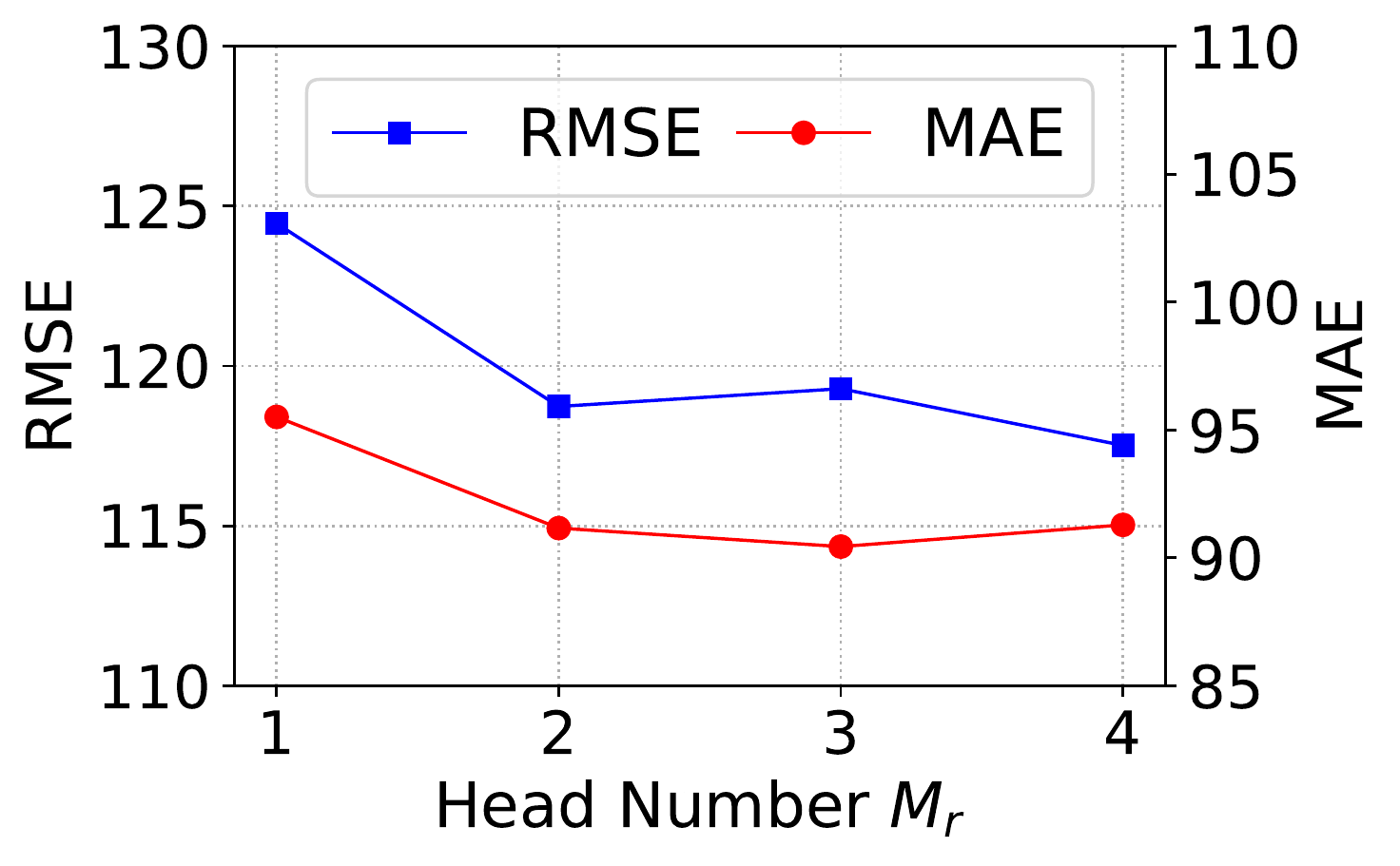}}
\vspace{-5mm}
\caption{Parameter analysis in CP task.}
\vspace{-6mm}
\label{fig_exp_param_nyc}
\end{figure}

\vspace{-1mm}
\subsubsection{Parameters Analysis.}
We further investigate the influence of several important hyper-parameters on the performance of \mymodel while keeping other parameters fixed. Figure \ref{fig_exp_param_nyc} presents the results in CP task, and other results are in Appendix \ref{addtional_results}.

\textbf{Number of spatial groups $n_s/n_r$.}
We first analyze the effects made by the number of partitioned sectors $n_s$ and rings $n_r$. By increasing $n_s/n_r$ to partition more subspaces, \mymodel can model more diverse spatial relations and further capture the diversity of spatial heterophily in a more fine-grained manner. However, over-dense partitions bring no further improvements but even slight performance declines. A possible explanation is that some subspaces contain too few neighbors to support its representation learning.

\textbf{Partition head number $M_s/M_r$.}
We also study the impact of head number $M_s/M_r$ in the multi-head partition strategy. It can be observed that, compared to the single partition ($M_s,M_r=1$), \mymodel using such a strategy ($M_s,M_r=2$) can evidently perform better,
thanks to the complementing role between two heads. When $M_s$ and $M_r$ continue to increase, the model generally gets fewer further improvements, and too many heads with additional redundancy may sometimes result in performance degradation. Thus, we recommend to set a small head number but larger than 1(e.g., 2), which is good enough while keeping efficiency. 

% \vspace{-0.5mm}
\section{Related Work}
Here we briefly review two related topics: GNNs for urban applications and  GNNs with heterophily.

% \textbf{GNNs for Urban Applications.}
% \vspace{-1mm}
% \paragraph{GNNs for Urban Applications.}
\textbf{GNNs for Urban Applications.}
As a powerful approach to representing relational data, GNN models \hide{like GCN \cite{kipf2016semi} and GAT \cite{velivckovic2017graph} }are widely adapted in recent studies to learn on urban graphs, and achieve remarkable performance in various \hide{urban }applications, including traffic \hide{flow }forecasting \cite{ xia20213dgcn, song2020spatial, fang2021spatial,yuan2021effective, rao2022fogs, wang2022event, NEURIPS2022_fanliu}, bike demand prediction \cite{li2022data}, \hide{road network representation \cite{wu2020learning},} region embedding \cite{wu2022multi, zhang2021multi}, regional economy prediction \cite{xu2020attentional} and special region discovery \cite{xiao2022contextual}. There are a few works that tend to encode the location information in GNNs' message passing process \cite{li2020competitive, chen2021points} in a special domain (POI relation prediction). But these methods fail to generalize to urban graphs with heterophily, which may significantly degrade their performance in other urban applications.

% \vspace{-1mm}
% \paragraph{GNNs with Heterophily.}
\textbf{GNNs with Heterophily.}
Our research is also related to the studies of graph heterophily. Here we only make a brief introduction to heterophilic GNNs and refer readers to a recent comprehensive survey \cite{zheng2022graph}.
Such approaches solve the heterophily problem basically in the following two ways.
The first branch is to reconstruct the homophilic neighborhood with similar nodes on the graph measured by different criteria. The used criteria \hide{for node similarity measuring }include the structural similarity defined by the distance in latent space \cite{pei2019geom} or degree sequence \cite{suresh2021breaking}, the difference of attention scores \cite{liu2021non}, cosine similarity of node attributes \cite{jin2021universal,jin2021node}, nodes' ability to mutually represent each other \cite{li2022finding} and so on.
However, as pointed out by \citet{he2022block}, these methods will damage the network topology and tamper with the original real-world dependencies on the urban graph.

Another branch tends to modify the GNN architecture to handle the difference information on heterophilic graphs, in contrast to the Laplacian smoothing \cite{wu2019simplifying} of typical GNNs, such as processing neighbors in different classes separately \cite{du2022gbk,dai2022label}, explicitly aggregating features from higher-order neighbors in each layer \cite{abu2019mixhop,zhu2020beyond}, 
allowing the high-frequency information by passing signed messages \cite{yan2021two, bo2021beyond, yang2021diverse, wu2022beyond, luan2021heterophily}, and combining outputs of each layer (including ego features) to also empower the GNNs with a high-pass ability \cite{chien2021adaptive, zhu2020beyond, chen2020simple}. 
However, most of these methods do not consider the diversity of dissimilarity distribution between the central node and different neighbors, especially with different spatial relations.

\vspace{-1mm}
\section{Conclusion}
In this paper, we studied the unique spatial heterophily of the urban graph and developed a spatial heterophily-aware graph neural network. We designed a spatial diversity score to uncover the diversity of heterophily at different spatial locations in the neighborhood, and showed the limitation of existing GNNs for handling diverse heterophily distributions on urban graphs. Further, motivated by the analysis that spatially close neighbors present a more similar mode of heterophily, we proposed a novel method, named \mymodel, which can group spatially close neighbors together and separately process each group with less diversity inside, to tackle the spatial heterophily in a divide-and-conquer way. Finally, extensive evaluations demonstrate the effectiveness of our approach.

\vspace{-1mm}
\begin{acks} 
This work is supported in part by Foshan HKUST Projects (FSUST21-FYTRI01A, FSUST21-FYTRI02A).
\end{acks}

\bibliographystyle{ACM-Reference-Format}
\balance
\bibliography{reference_abb}

\clearpage
\appendix
\nobalance
\section{Appendix}
\label{apdx}
\subsection{Datasets Construction}
\label{apdx_dataset}
In this section, we make a supplementary explanation of the construction of three real-world datasets in three tasks.

\vspace{-1mm}
\subsubsection{Commercial Activeness Prediction (CAP)}
Following many previous studies (e.g. \cite{liang2021fine, wang2022event}) that divide a city into grids for various urban analyses, we also divide Shenzhen city into a set of $128m \times 128m$ grids, which are regarded as the regions in this task. 
We construct two groups of region features for commercial activeness prediction, which are POI features and satellite image features. 

% \paragraph{\textit{A. POI Features.}}
\textit{\textbf{A. POI Features.}}
We construct three types of POI features from the POI data, to reflect a region's functionality, which is highly correlated with commercial activeness \cite{xi2022beyond}.

\textit{(a) Category Distribution.}
For each region, we compute the ratios of different categories of POIs, to obtain the category distribution histogram as a feature vector. 
The total number of POIs is also appended in this vector. In our experiments, we consider the following 23 categories: Food Service, Hotel, Shopping Place, Life Service, Beauty Industry, Scenic Spot, Leisure and Entertainment, Sports and Fitness, Education, Cultural Media, Medicine, Auto Service, Transportation Facility, Financial Service, Real Estate, Company, Government Apparatus, Entrance and Exit, Topographical Object, Road, Railway, Greenland, and Bus Route. 
To include more surrounding information, we also compute the distribution in the $3 \times 3$ regions centered by the given region.

\textit{(b) POI Radius.}
This feature is defined as the shortest distance between a certain type of POIs and the region, such as hospital radius. 
The distance is discretized into different buckets ($<0.5km$, $0.5\sim1.5km$, $1.5\sim3km$ and $>3km$). 
We calculated 15 kinds of radius in total: Hospital, Clinic, College, School, Bus Stop, Subway Station, Airport, Train Station, Coach Station, Shopping Mall, Supermarket, Market, Shop, Police Station, and Scenic Spot. 

\textit{(c) Perfect Degree of Living Facility.} 
It's a binary index that reflects whether residents can satisfy their daily consumption demands around the region. It will be assigned one if there are all the following types of facilities within $1km$ of the current region, which are Medical Service, Shopping Place, Sports Venue, Education Service, Food Service, Financial Service, Communication Service, Public Security Organ and Transportation Facility. We select these types of POIs according to the official document 
\footnote{http://www.mohurd.gov.cn/wjfb/201811/W02018113004480.pdf}.

% \paragraph{B. Satellite Image Features.}
\textit{\textbf{B. Satellite Image Features.}}
The satellite image data are 3-channel $256 \times 256$ RGB images with 0.5 spatial resolution. 
We use the pre-trained VGG16 \cite{simonyan2014very} with the last two dense layers removed, to extract semantic features from the satellite image of each region.

\vspace{-1mm}
\subsubsection{Crime Prediction (CP)}
We provide more details about the Manhattan dataset.
As introduced in \cite{wu2022multi,zhang2021multi}:
(1) \textit{Census block data} contains the boundaries of 180 regions split by streets in Manhattan. (2) \textit{Taxi trip data} records around 10 million taxi trips during one month among regions. (3) \textit{POI data} includes information of nearly 20 thousand POIs, which can be divided into 14 categories. (4) \textit{Crime data} consists of more than 35 thousand crime records during one year in these 180 regions.
We construct 14-dimensional POI category distribution region features similar to CAP task.

\vspace{-1mm}
\subsubsection{Dangerous Road Section Detection (DRSD)}
We further introduce the road data and traffic accident record data used to build the Los Angeles dataset.
The road data describes the location and connecting relationship of road sections in a city. It is used to build the graph of road networks, where nodes are road sections, and edges are added between connected sections. 
For each node, we apply Deepwalk \cite{perozzi2014deepwalk} to generate the embedding as node features.
As for the accident data, it records the occurrence time and location of nearly 20 thousand car accidents in December 2021.

\begin{table*}[t]
\renewcommand\arraystretch{0.8}
\small
\caption{Efficiency comparison in the commercial activeness prediction task.}
\label{table-time}
\vspace{-3mm}
\centering
% \begin{tabular}{c|p{1.3cm}<{\centering}|p{1.3cm}<{\centering}|p{1.3cm}<{\centering}|p{1.3cm}<{\centering}}
\resizebox{0.9\textwidth}{!}{
\begin{tabular}{c|c|c|c|c|c|c|c|c|c|c|c}
    \toprule
    & GCN & GAT & Mixhop & FAGCN & NLGCN & GBKGNN & GPRGNN & PRIM & SAGNN & KnowCL & \mymodel \\
    % \cline{1-12}
    % \rule{0pt}{10pt}
    \midrule
    Training time (s / epoch) & 0.054 & 0.125 & 0.113 & 0.105 & 0.070 & 0.222 & 0.057 & 0.172 & 0.474 & 0.017 & 0.392 \\
    Inference time (s / instance) & 0.003 & 0.004 & 0.033 & 0.005 & 0.031 & 0.005 & 0.003 & 0.006 & 0.008 & 0.001 & 0.018 \\
    \bottomrule
\end{tabular}
}
\vspace{-3mm}
\end{table*}
\vspace{-2mm}
\subsection{Implementations}
% \subsubsection{Parameter Settings.}
We first introduce some basic settings (e.g., input and output layers) for three tasks. For CAP task, since the input POI and satellite image features are from different modalities, we apply two dense layers with 64 hidden units to transform them into the same space, and concatenate them together to obtain the initial representation (i.e., $\bm{h}_{i}^{(0)}$) of each region. 
For CP task, the 14 POI category features are directly used as the initial representation. 
For DRSD task, the node representation is initialized by the 64-$d$ embedding vector learned by Deepwalk algorithm \cite{perozzi2014deepwalk}, where we set the walk length as 10, the number of random walk as 80, and the window size as 5.
As for the output layer, we adopt Linear Regression to get the final prediction in CAP and CP task. For the node classification in DRSD task, we use Logistic Regression to obtain the output probability.
Moreover, for all three tasks, we use Adam optimizer for model training with the learning rate set to 0.001. 

Next, we introduce other hyper-parameter settings in \mymodel. For rotation-scaling spatial aggregation module, we set the head number $M_s=M_r=2$ to divide neighbors from both the direction and distance view. Specifically, we partition $n_s=4$ sectors under each head and the rotation angle is set to 45 degrees for all three tasks. 
In the distance view, we set different ring numbers in three tasks.
For CAP task and CP task, we partition $n_r=3$ rings per head, and the interval of each distance bucket is set to $1.5km$ (under the head $m=1$) and $2.5km$ (head $m=2$), while for DRSD task, we set $n_r=2$ with $100m$ (head $m=1$) and $200m$ (head $m=2$) as bucket intervals. The hidden size is set to 32 for all the tasks.
For heterophily-sensitive spatial interaction module, the hidden size of the two dense layers used to fuse two spatial views are set to 32, 128, and 32 for CAP, CP, and DRSD task, respectively.
We apply one \mymodel layer between the input and output layer to learn the node representation based on the urban graph.

\vspace{-2mm}
\subsection{Baseline Descriptions}
\label{baseline_describe}
\begin{itemize}[leftmargin=*]
    \setlength{\itemsep}{2pt}
    \item \textbf{GCN} \cite{kipf2016semi} is a classic message passing homophilic GNN model.
    
    \item \textbf{GAT} \cite{velivckovic2018graph} uses self-attention mechanism in the feature aggregation. It also implicitly assumes that the graph is homophilic.
    
    \item \textbf{Mixhop} \cite{abu2019mixhop} is a GNN model for heterophilic graphs. It directly aggregates features from higher-order neighbors, which can identify feature differences among neighbors.
    
    \item \textbf{FAGCN} \cite{bo2021beyond} is a heterophily GNN that extends the message from neighbors to be signed value. Its aggregation process can capture the high-frequency signals in the neighborhood. 
    
    \item \textbf{NLGCN} \cite{liu2021non}
    tackles the heterophily by rebuilding the homophilic neighborhood for similar nodes. The similarity of two nodes is measured by their attention scores to a calibration vector.
    
    \item \textbf{GPRGNN} \cite{chien2021adaptive} combines intermediate representations from each GNN layer (including ego features). It achieves remarkable performance on general heterophilic graphs.

    \item \textbf{GBKGNN}
    \cite{du2022gbk} tackles the graph heterophily problem by applying two different feature transformations for homophily node pairs and heterophily node pairs, respectively. 
    
    \item \textbf{SAGNN} \cite{li2020competitive} includes the spatial information of neighbors in feature aggregation. It considers neighbors' direction by splitting sectors, and processes their relative position by dividing grids.
    
    \item \textbf{PRIM} \cite{chen2021points} is a GNN-based framework working on POI graphs, with the consideration of geographical influence. When aggregating neighbor features, it modifies the attention score with a spatial distance score calculated by radial basis function (RBF).

    \item \textbf{KnowCL} \cite{liu2023knowledge} is specifically designed for socioeconomic prediction. It uses GCN encoder to extract multiple knowledge from an urban knowledge graph, and infuse it to region's visual representation through cross-modality contrastive learning.

    \item \textbf{NNCCRF} \cite{yi2019neural} is a crime prediction method. It integrates Continuous Conditional Random Field (CCRF) with neural network to model region interactions for crime prediction.

    \item \textbf{RFN} \cite{jepsen2020relational} is an improved GCN designed for prediction tasks on road networks. It simultaneously captures two interdependent views of relations, between both intersections and road segments.
    
\end{itemize}
% \vspace{-2mm}
\subsection{Additional Experimental Results}
\label{addtional_results}

\textbf{Efficiency.} We also conducted an experiment in CAP task to evaluate the efficiency of \mymodel. We compare \mymodel and all baselines in this task, in terms of the training time per epoch and the inference time of an instance. 
As shown in Table \ref{table-time}, the efficiency of \mymodel is affordable (0.392 seconds for training one epoch, and 0.018 seconds for inferring one instance), since our model can handle spatial heterophily on the urban graph, and achieves significant performance improvements over all baselines.

\begin{figure}[t]
\centering
\vspace{-1mm}
\subfigure{
    \includegraphics[width=0.43\columnwidth]{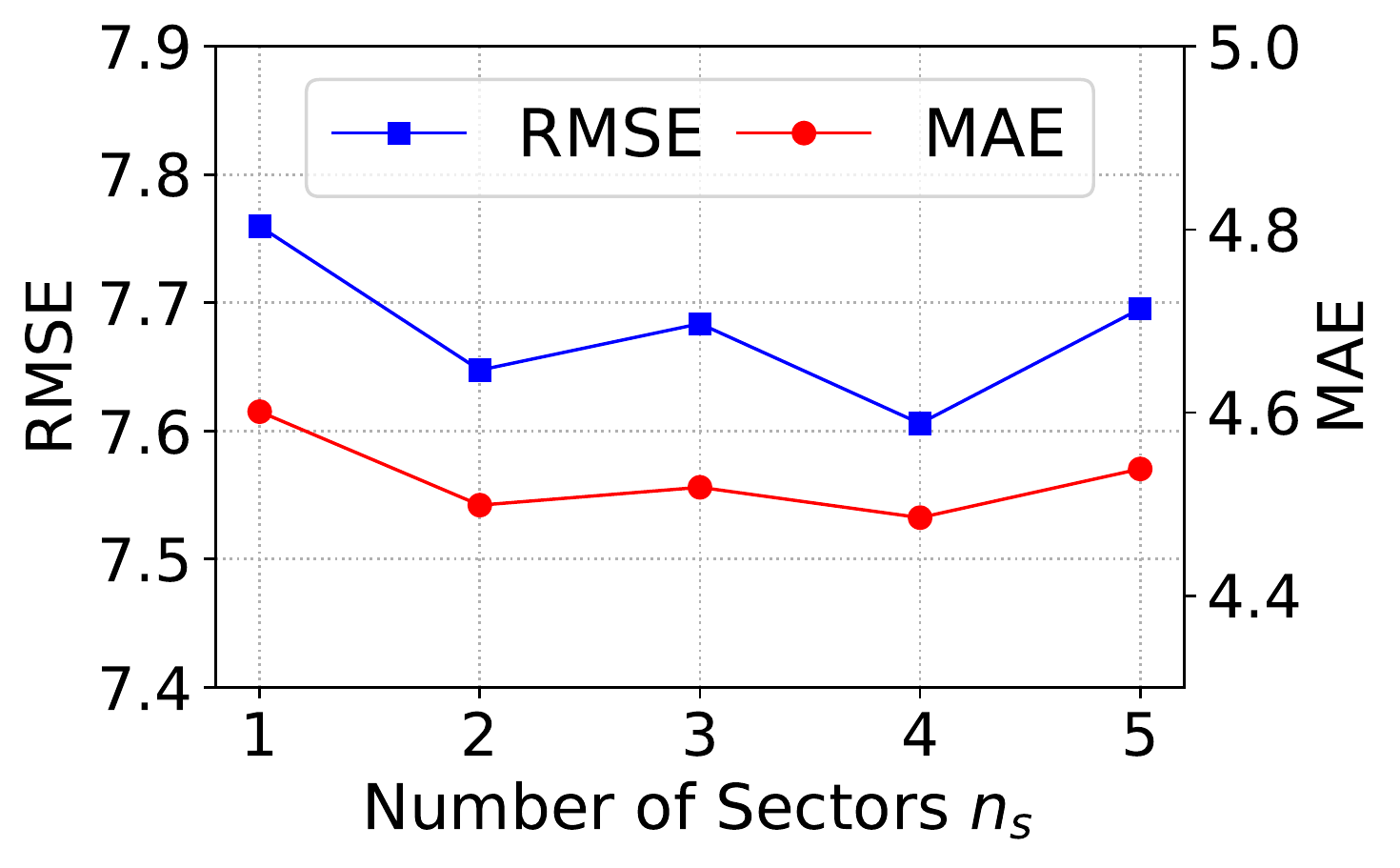}}
\subfigure{
    \includegraphics[width=0.43\columnwidth]{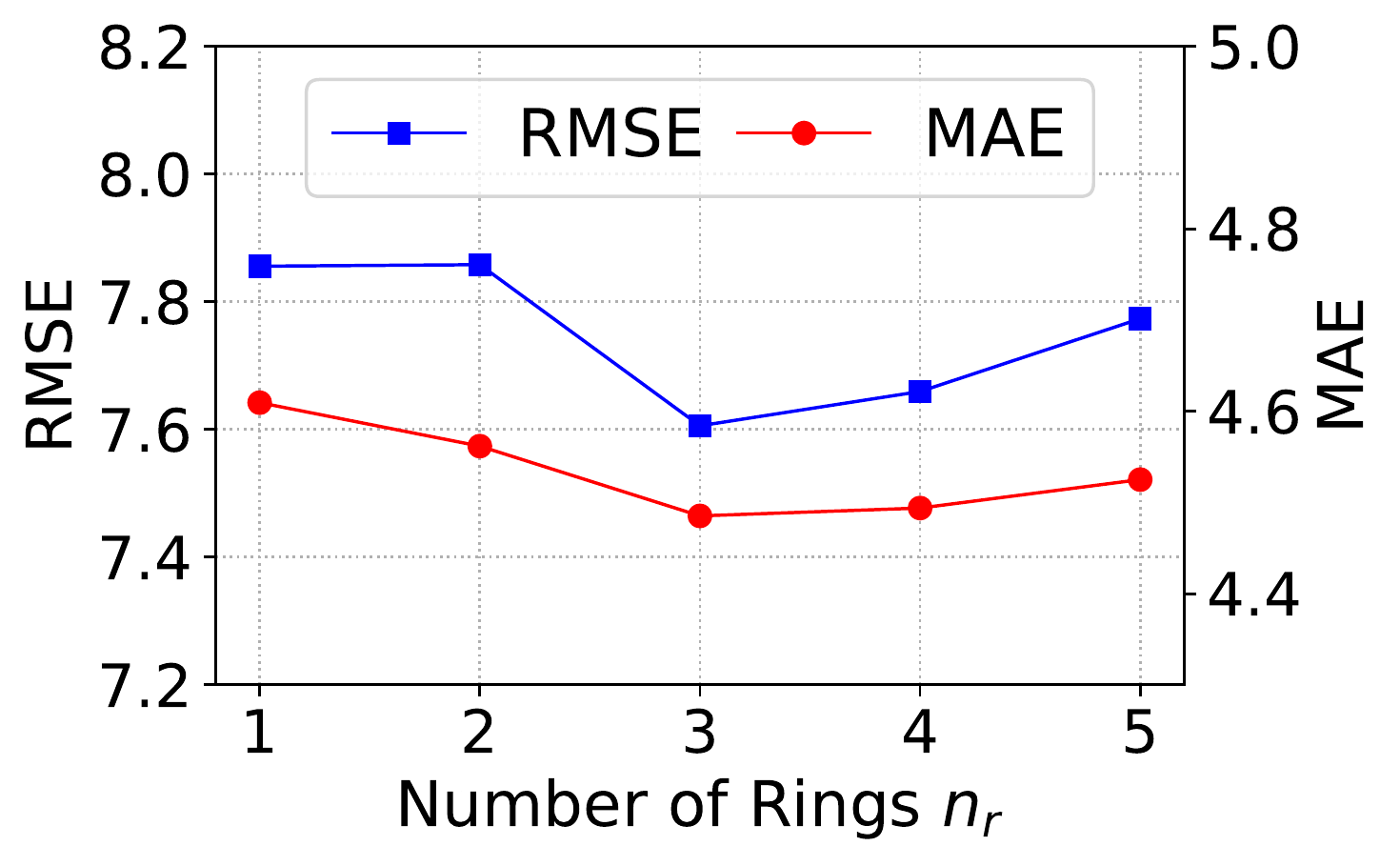}} \\
    \vspace{-4mm}
\subfigure{
    \includegraphics[width=0.43\columnwidth]{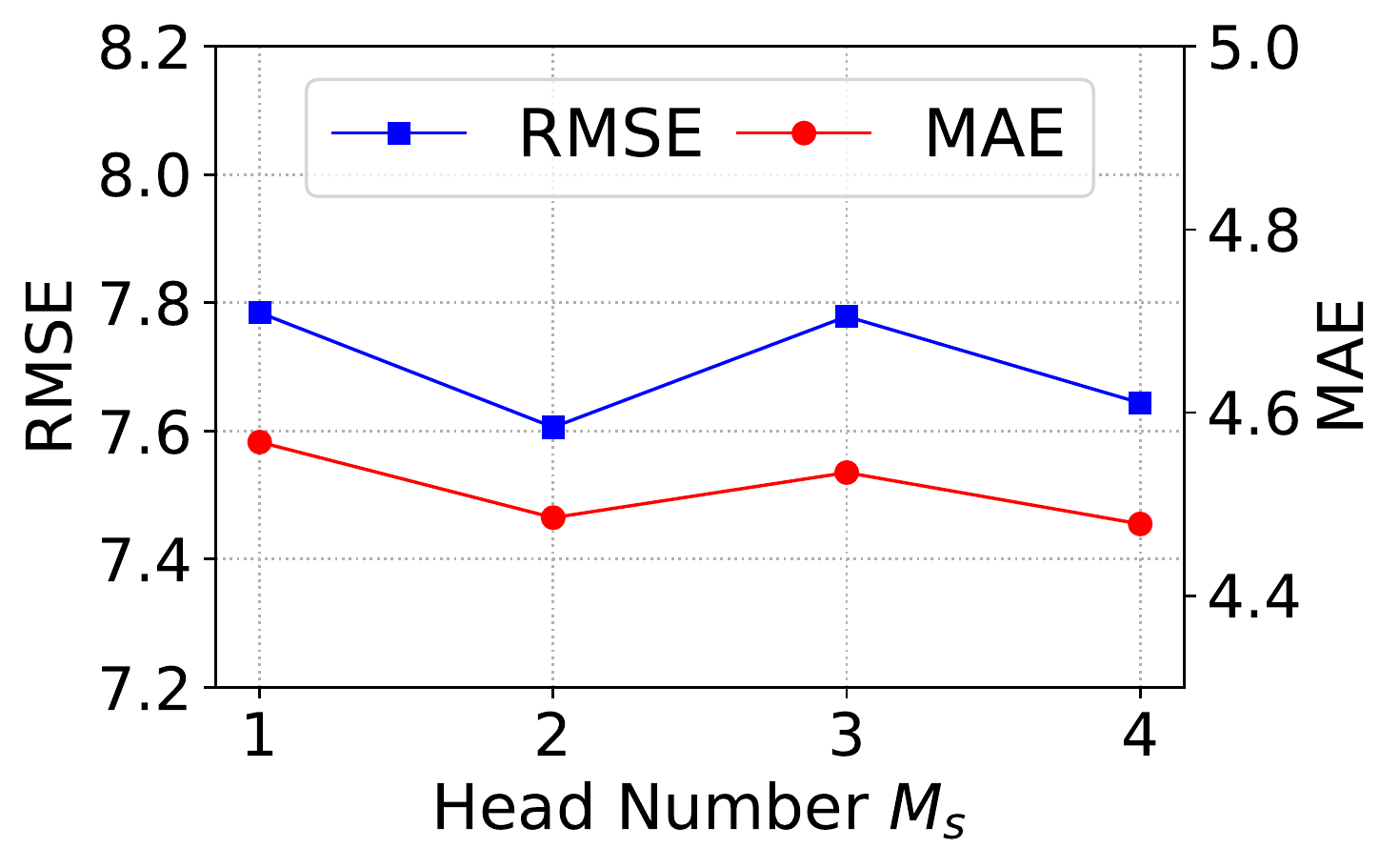}}
\subfigure{
    \includegraphics[width=0.43\columnwidth]{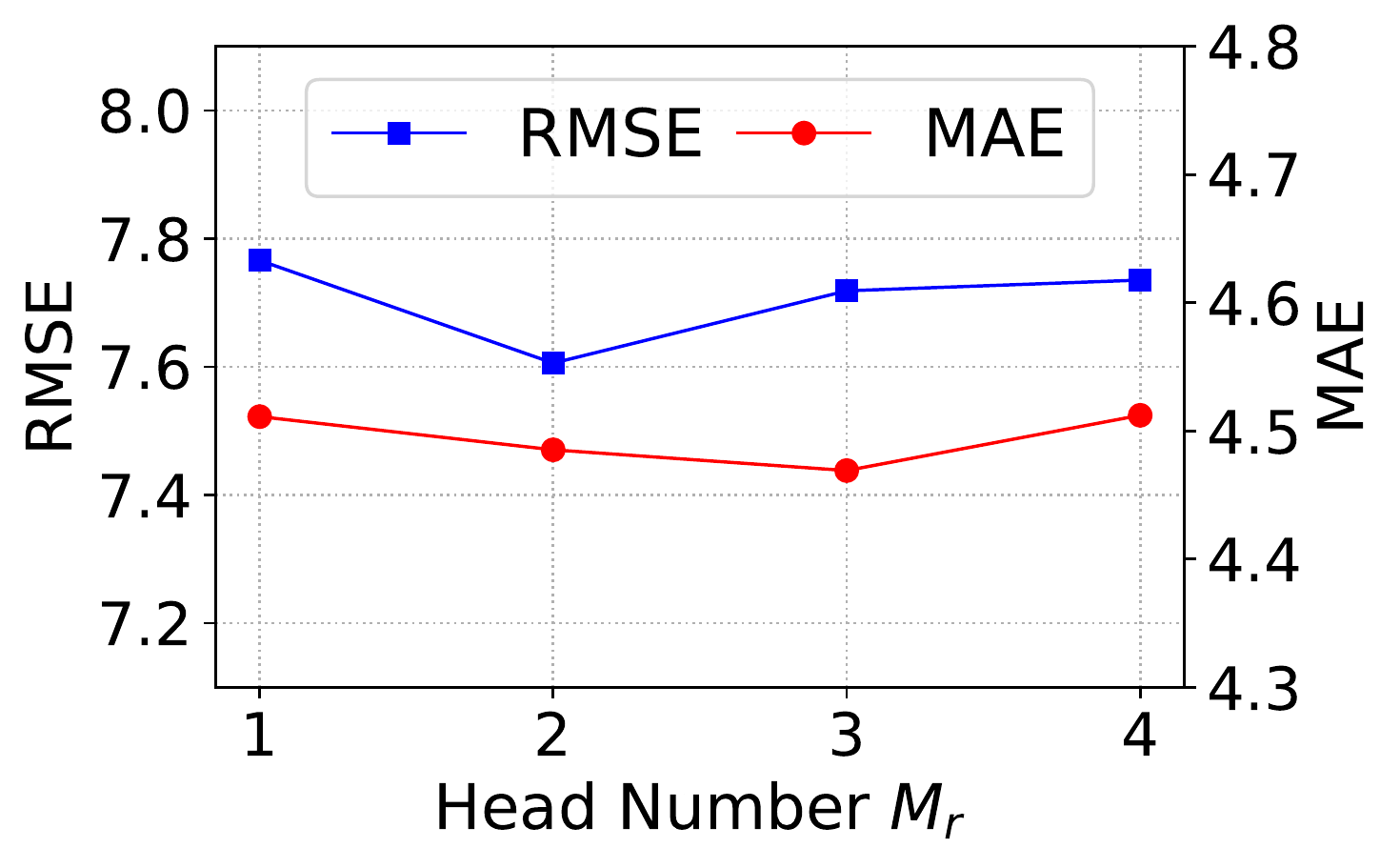}}
\vspace{-5.5mm}
\caption{Parameter analysis in CAP task.}
\vspace{-4mm}
\label{fig_exp_param_cmt}
\end{figure}

\begin{figure}[t]
\centering
% \vspace{-2mm}
\subfigure{
    \includegraphics[width=0.45\columnwidth]{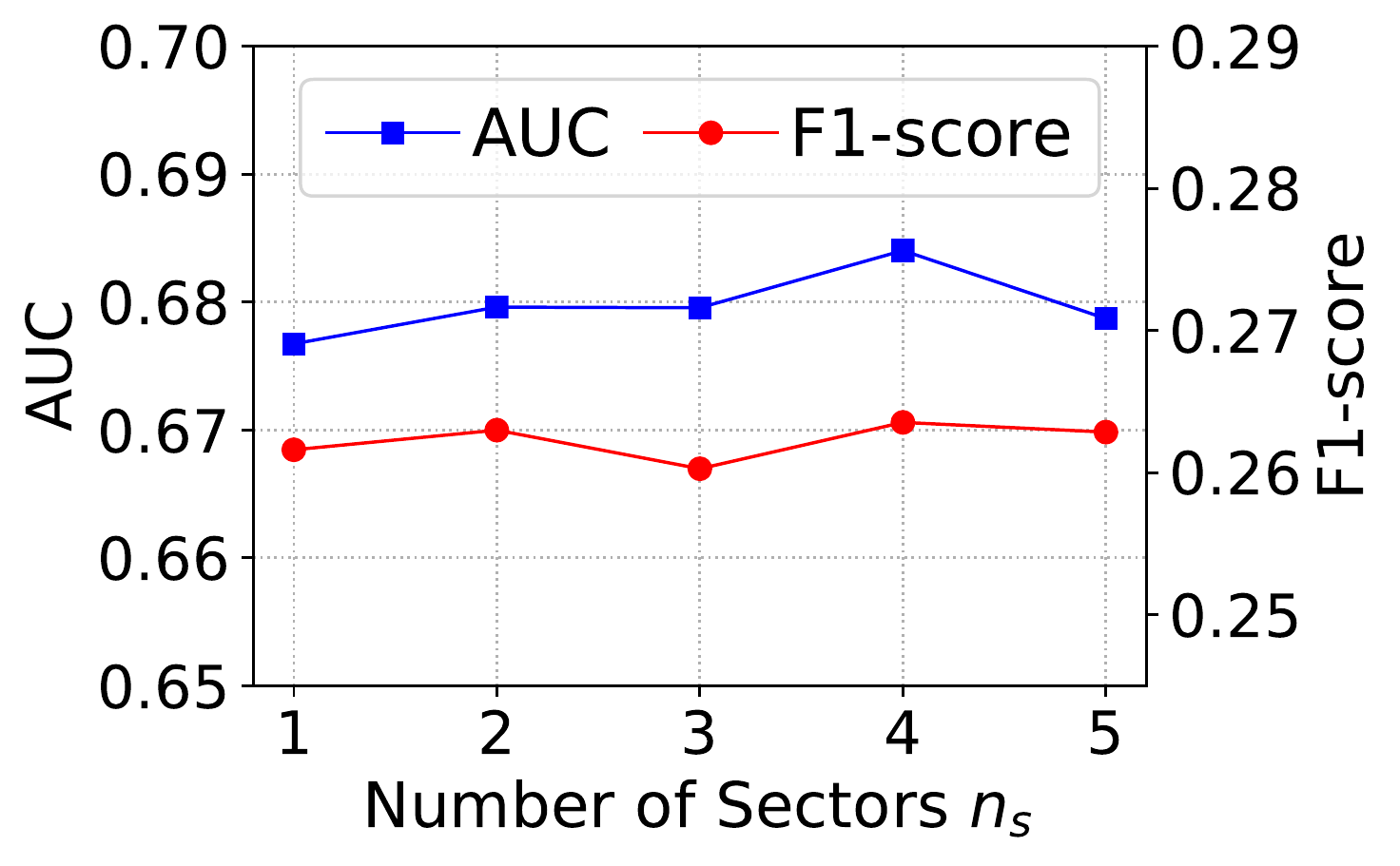}}
\subfigure{
    \includegraphics[width=0.45\columnwidth]{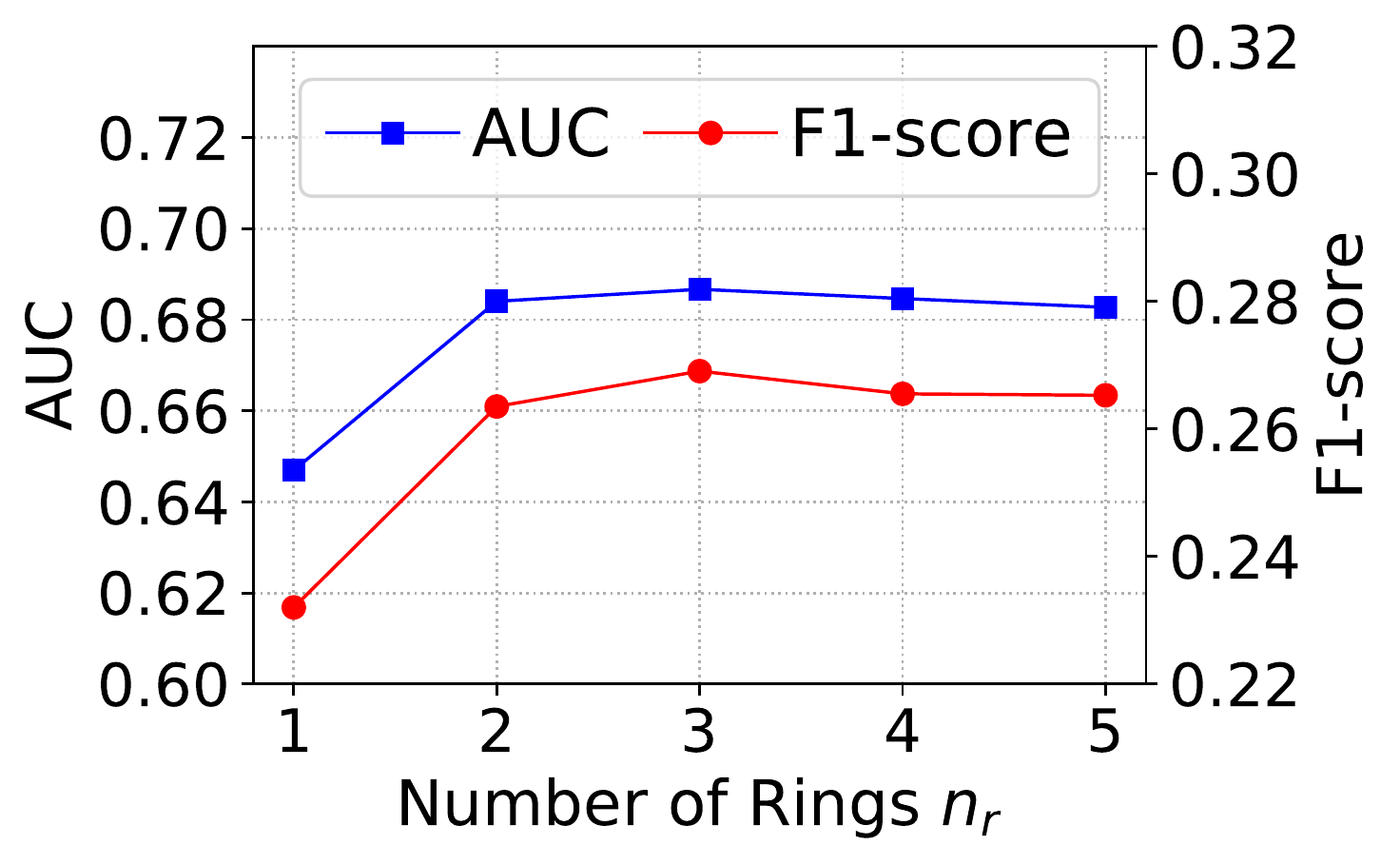}} \\
    \vspace{-4mm}
\subfigure{
    \includegraphics[width=0.45\columnwidth]{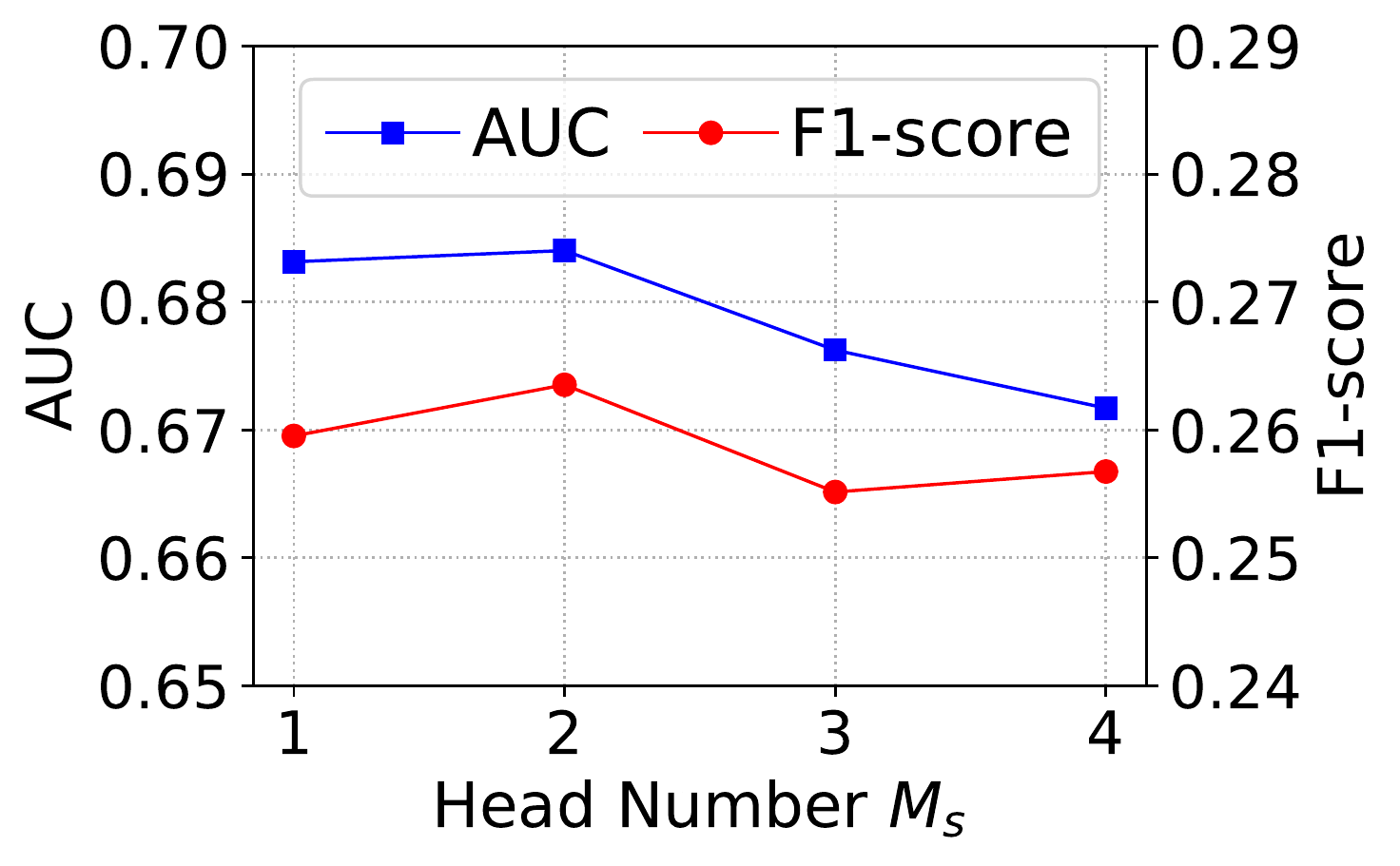}}
\subfigure{
    \includegraphics[width=0.45\columnwidth]{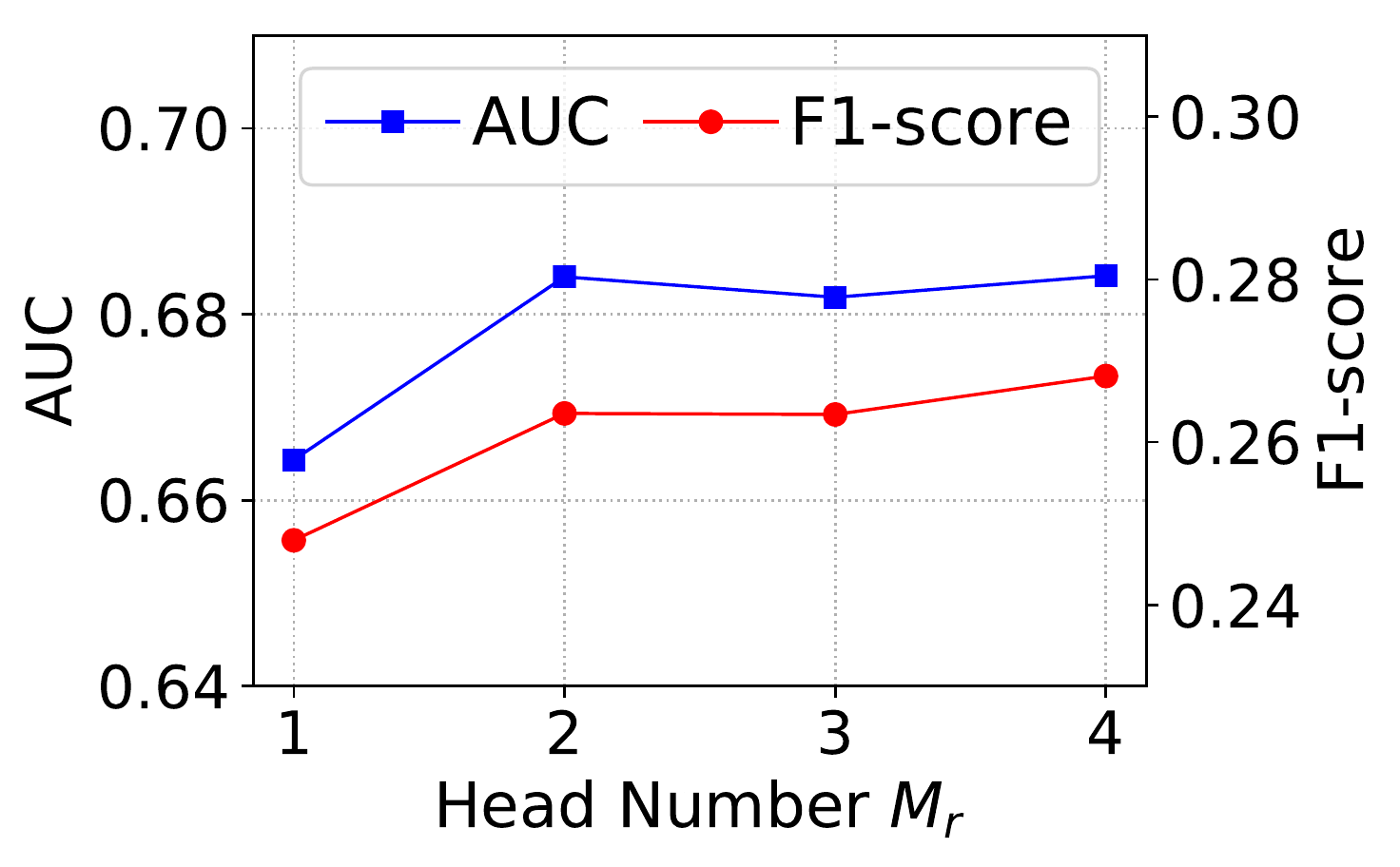}}
\vspace{-5.5mm}
\caption{Parameter analysis in DRSD task.}
\vspace{-1mm}
\label{fig_exp_param_drsd}
\end{figure}

\textbf{Parameter Analysis.} The results of parameter analysis in CAP and DRSD task are shown in Figure \ref{fig_exp_param_cmt} and \ref{fig_exp_param_drsd}.

% \vspace{-2mm}
\subsection{Complexity Analysis}
We further analyze the complexity of \mymodel.
First, in the rotation-scaling spatial aggregation module, the computation cost comes from the spatial-aware aggregation process. Its complexity is:
\begin{equation}
    \mathcal{O} ((M_s n_s + M_r n_r) |\mathcal{V}| d^2 + (M_s+M_r) |\mathcal{E}| d),
    \nonumber
\end{equation}
where $|\mathcal{V}|$ and $|\mathcal{E}|$ denote the number of nodes and edges, and $d$ denotes the node feature dimension. $n_s$ and $n_r$ denote the number of sectors and rings. $M_s$ and $M_r$ are head numbers of the multi-head sector partition in direction view and ring partition in distance view.
To explain, in the direction view, the term $\mathcal{O}(M_s n_s |\mathcal{V}| d^2)$ is the cost of feature transformation for nodes in $n_s$ different sectors, under $M_s$ heads of sector partitions. And $\mathcal{O} (M_s |\mathcal{E}| d)$ corresponds to the cost of the feature aggregation process. Similarly, the part $\mathcal{O} (M_r n_r |\mathcal{V}| d^2 + M_r |\mathcal{E}| d)$ is for the distance view. 
% In practice, there is no need to set a large number of sectors ($n_s$) and rings ($n_r$), as well as the partition heads ($M_s$ and $M_r$). In our experiments, we set $n_s=4$, $n_r=2,3$ and $M_s=M_r=2$.

In the heterophily-sensitive spatial interaction module, the complexity of commonality and discrepancy kernel are both:
\begin{equation}
    \mathcal{O} ((M_s n_s + M_r n_r) |\mathcal{V}| d^2 + (M_s n_s^2 + M_r n_r^2) |\mathcal{V}| d),
    \nonumber
\end{equation}
where the first term is to extract the common knowledge / difference information, and the second term is the cost to compute the commonality / dissimilarity degree among sectors and among rings. Then, the computation cost of attentive component selection is also $\mathcal{O} ((M_s n_s + M_r n_r) |\mathcal{V}| d^2)$. 
Overall, the total complexity of \mymodel is the combination of the above two modules:
\begin{equation}
    \mathcal{O}((M_s n_s + M_r n_r) |\mathcal{V}| d^2 + (M_s n_s^2 + M_r n_r^2) |\mathcal{V}| d + (M_s + M_r) |\mathcal{E}| d).
    \nonumber
\end{equation}

\renewcommand\arraystretch{1.}
\begin{table}[t]
    \footnotesize
\caption{Mathematical notations.}
\vspace{-5mm}
\label{table-symbol}
\resizebox{0.47\textwidth}{!}{
\begin{tabular}{cl}
    \midrule
    \midrule
    Notation	&	Description	\\
    \midrule
    % $\mathcal{G}(\mathcal{V}, \mathcal{E}, X)$ & The urban graph with node set $\mathcal{V}$, edge set $\mathcal{E}$, node attributes $X$ \\
    $\mathcal{G}$ & The urban graph \\
    $\mathcal{V}, \mathcal{E}, \bm{X}$ & The node set, edge set and node attributes of $\mathcal{G}$ \\
    $\mathcal{N}(v_i)$ & The neighborhood of node $v_i$ on $\mathcal{G}$ \\
    $\bm{h}_{i}$ & The representation of node $v_i$ \\
    $s_k$, $r_k$	& The partitioned direction sector and distance ring \\
    $\mathcal{N}_{s_k}(v_i)$ & The direction-aware neighborhood of node $v_i$ \\
    $\mathcal{N}_{r_k}(v_i)$ & The distance-aware neighborhood of node $v_i$ \\
    $\bm{z}_{i, s_k}$ & The representation of sector $s_k$ around node $v_i$ \\
    $\bm{z}_{i, r_k}$ & The representation of ring $r_k$ around node $v_i$ \\
    \midrule
    \midrule
\end{tabular}
}
\vspace{-3mm}
\end{table}

\end{document}